%% file: main.tex
\documentclass[10pt]{article} 

\input{math_commands.tex}

\usepackage{hyperref}
\usepackage{url}
\usepackage{placeins}

\usepackage{tcolorbox}
\usepackage{todonotes}
\makeatletter
\define@key{todonotes}{EK}[]{%
    \setkeys{todonotes}{author=EK,color=green}}%
\definecolor{lightred}{RGB}{255,71,76}
\define@key{todonotes}{WL}[]{%
    \setkeys{todonotes}{author=WL,color=orange}}%
\define@key{todonotes}{WX}[]{%
    \setkeys{todonotes}{author=WX,color=cyan}}%
\define@key{todonotes}{SS}[]{%
    \setkeys{todonotes}{author=SS,color=yellow}}%
\usepackage{multirow} 
\usepackage{booktabs}
\usepackage{amsmath,amsfonts,amssymb,amsthm}
\usepackage{enumitem}
\usepackage{cleveref}
\usepackage{array}
\usepackage{wrapfig}

\usepackage[accepted]{tmlr}

\title{Tackling the Abstraction and Reasoning Corpus \\ with Vision Transformers: the Importance of \\ 2D Representation, Positions, and Objects}

\author{\name Wenhao Li \email chriswenhao.li@mail.utoronto.ca \\
      \addr Department of Mechanical \& Industrial Engineering, University of Toronto
      \AND
      \name Yudong Xu \email wil.xu@mail.utoronto.ca \\
      \addr Department of Mechanical \& Industrial Engineering, University of Toronto 
      \AND
      \name Elias B.~Khalil \email khalil@mie.utoronto.ca\\
      \addr Department of Mechanical \& Industrial Engineering, University of Toronto \\
      Scale AI Research Chair in Data-Driven Algorithms for Modern Supply Chains
      \AND
      \name Scott Sanner \email ssanner@mie.utoronto.ca\\
      \addr Department of Mechanical \& Industrial Engineering, University of Toronto\\
      Vector Institute for Artificial Intelligence
}

\def\month{05}  
\def\year{2025} 
\def\openreview{\url{https://openreview.net/forum?id=Al72Fp0rCg}} 

\newcommand{\method}{\textsc{ViTARC}}

\begin{document}

\maketitle

\begin{abstract}
The Abstraction and Reasoning Corpus (ARC) is a popular benchmark focused on \textit{visual reasoning} in the evaluation of Artificial Intelligence systems. In its original framing, an ARC task requires solving a program synthesis problem over small 2D images using a few input-output training pairs. In this work, we adopt the recently popular \textit{data-driven} approach to the ARC and ask whether a Vision Transformer (ViT) can learn the implicit mapping, from input image to output image, that underlies the task.  We show that a ViT---otherwise a state-of-the-art model for images---fails dramatically on most ARC tasks even when trained on one million examples per task. This points to an inherent representational deficiency of the ViT architecture that makes it incapable of uncovering the simple structured mappings underlying the ARC tasks. Building on these insights, we propose~\method{}, a ViT-style architecture that unlocks some of the visual reasoning capabilities required by the ARC.  Specifically, we use a pixel-level input representation, design a spatially-aware tokenization scheme, and introduce a novel object-based positional encoding that leverages automatic segmentation, among other enhancements. 
Our task-specific~\method{} models achieve a test solve rate close to 100\% on more than half of the $400$ public ARC tasks strictly through supervised learning from input-output grids.
This calls attention to the importance of imbuing the powerful (Vision) Transformer with the correct inductive biases for abstract visual reasoning that are critical even when the training data is plentiful and the mapping is noise-free. Hence, \method{} provides a strong foundation for future research in visual reasoning using transformer-based architectures.\textsuperscript{\ref{fn:coderepo}}

\end{abstract}


\footnotetext[1]{\label{fn:coderepo}\textbf{Code} available at \url{https://github.com/khalil-research/ViTARC}}

%




\section{Introduction}

\begin{figure}[t]
    \centering
    \includegraphics[width=0.85\linewidth]{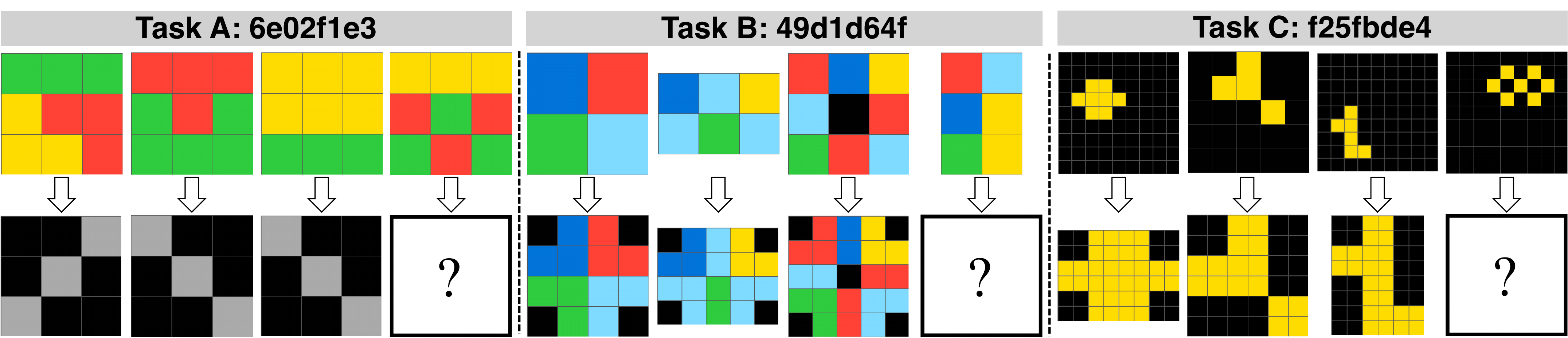}
    \caption{\textbf{Three example ARC tasks.} For each task, the first columns contain example input-output pairs from the ``training'' instances, and the last column contains the ``test'' instance. The goal is to use the training instances to solve the test instance. The Vanilla ViT setup (\Cref{sec:defaultVit}) was only able to solve Task A\textsuperscript{\ref{fn:taskA}}. Our ViTARC-VT (\Cref{sec:VisualTokens}) was able to solve Task A and B but still failed at Task C. Our final model ViTARC (\Cref{sec:PositionalEnhancements}) achieves near 100\% accuracy on all three tasks.}
    \label{fig:arc-examples}
\end{figure}

Developing systems that are capable of performing abstract reasoning has been a long-standing challenge in Artificial Intelligence (AI). Abstract Visual Reasoning (AVR) tasks require AI models to discern patterns and underlying rules within visual content, offering a rigorous test for evaluating AI systems. Unlike other visual reasoning benchmarks such as Visual Question Answering (VQA)~\citep{vqa} and Visual Commonsense Reasoning (VCR)~\citep{figureqa} that rely on natural language inputs or knowledge of real-world physical properties, AVR tasks do not include any text or background knowledge. Instead, they focus purely on visual abstraction and pattern recognition~\citep{malkinski2023review}.
One prominent example of AVR is the Abstraction and Reasoning Corpus (ARC)~\citep{chollet2019measure}, which is designed to evaluate an AI's capacity for generalization in abstract reasoning. Each ARC task involves transforming input grids into output grids by identifying a hidden 
mapping often requiring significant reasoning beyond mere pattern matching (cf. \Cref{fig:arc-examples}). While the ARC's original setting is one of few-shot learning, there has been recent interest in studying the ARC in a data-rich setting where task-specific input-output samples can be generated~\citep{hodel_rearc_2024}, allowing for the evaluation of deep learning-based solutions.

\begin{figure}[t]
    \centering
    \includegraphics[width=\linewidth]{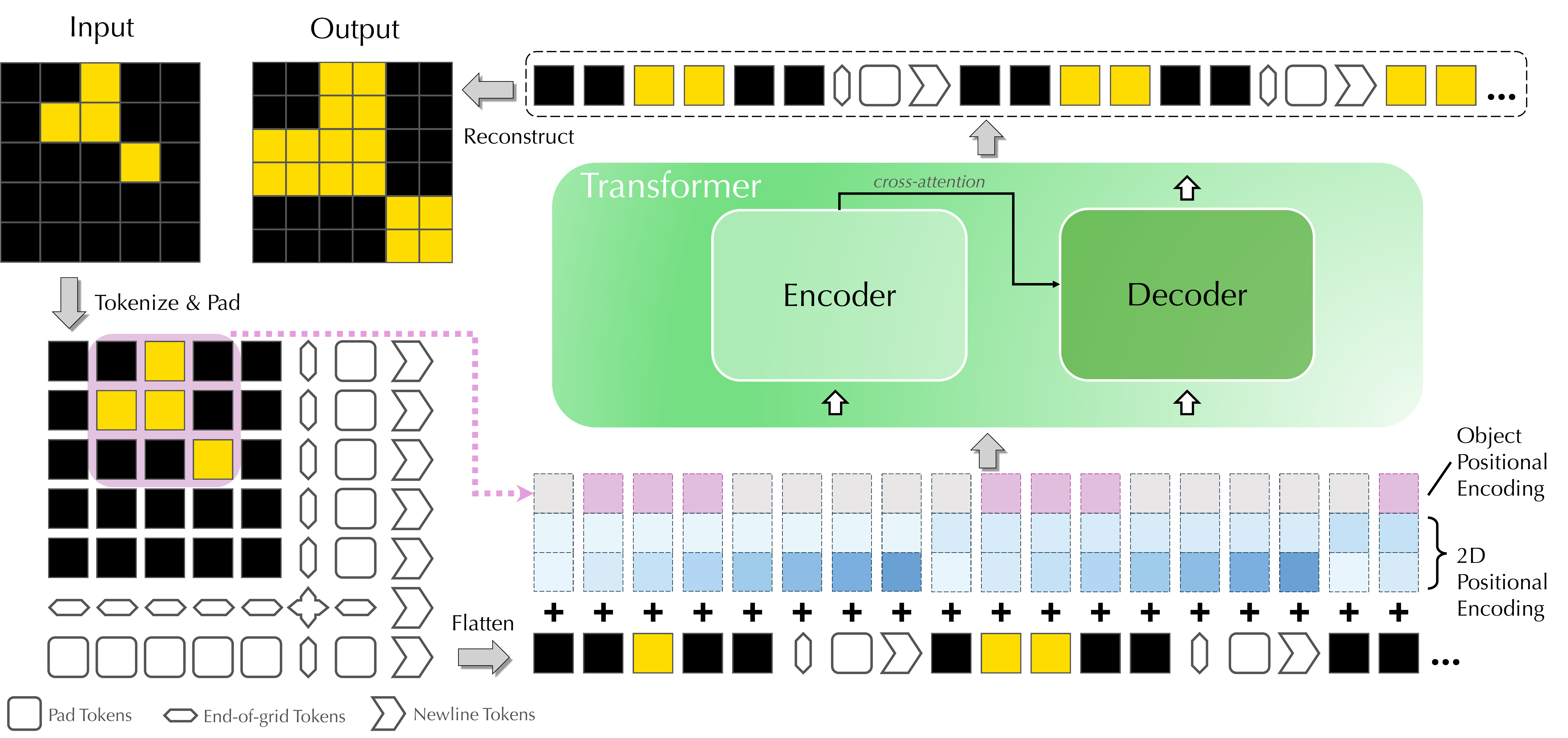}
    \caption{\textbf{Overview of our ViTARC framework contribution.} An ARC input image is first tokenized into pixels and padded with visual tokens including end-of-grid tokens that mark the end of the image grid, newline tokens that indicate the end of one row, and pad tokens which are used to pad the image into a fixed maximum size (not drawn in full to maintain clarity). 2D Positional Encodings and Object Positional Encodings are then added to each token before being passed into the transformer. The output tokens are reconstructed into a valid two-dimensional grid.}
    \label{fig:overview}
\end{figure}

\begin{figure}[htbp!]
\begin{center}
\includegraphics[width=0.90\linewidth]{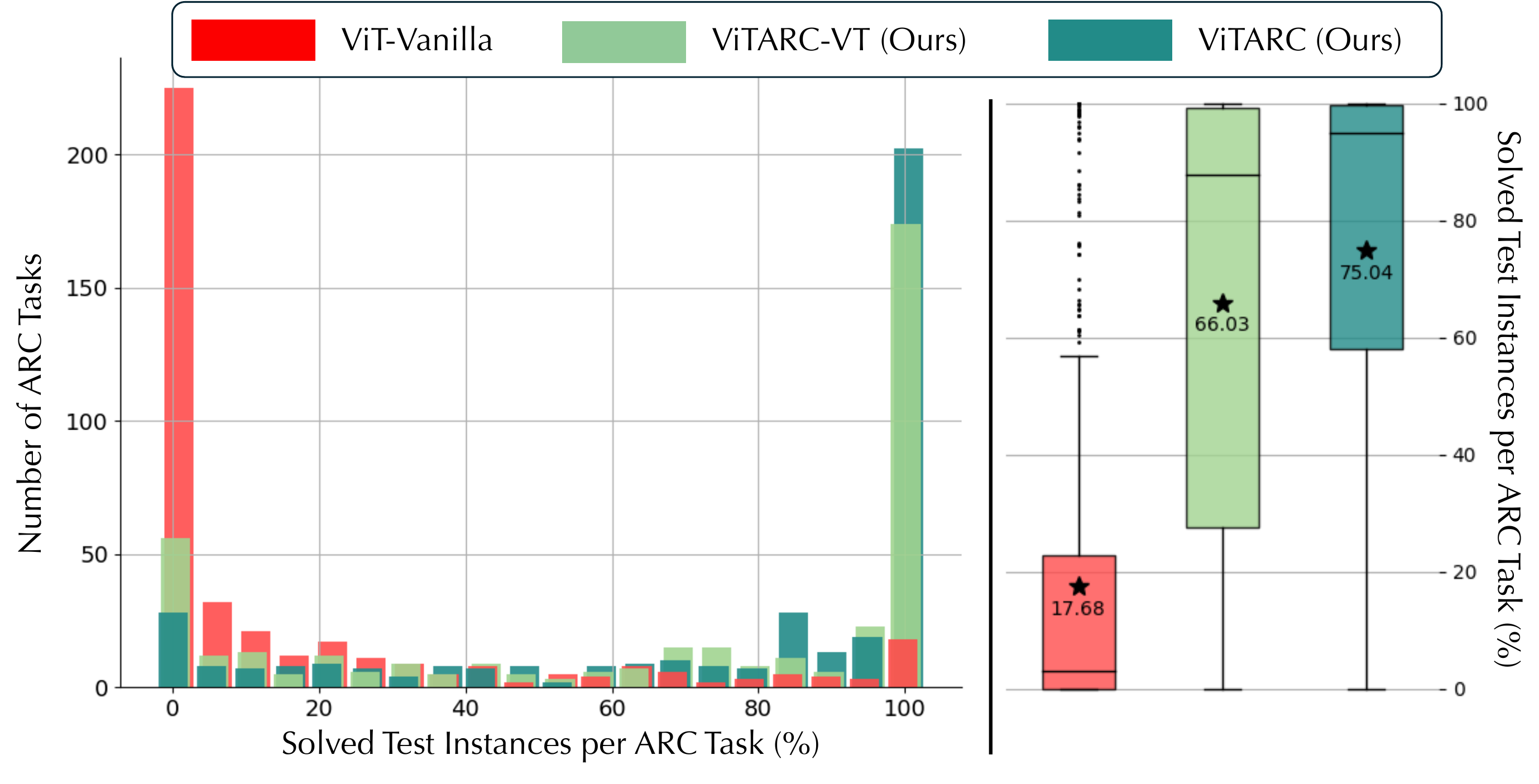}
\end{center}
\caption{\textbf{Model performances on 400 ARC tasks.} Three models are shown: ViT-Vanilla (red) represents the vanilla vision transformer setup (cf.~\Cref{sec:defaultVit}); ViTARC-VT (light green) and ViTARC (dark green) represent the variants of our framework introduced in Sections \ref{sec:VisualTokens} and \ref{sec:PositionalEnhancements}, respectively. (Left) Distribution of Solve Rates: The horizontal axis shows the solve rate (percentage of test instances that are solved correctly) on 1000 test instances per task. The vertical axis displays the number of tasks at each solve rate level. (Right) Distribution Statistics: The stars and corresponding values are the overall solve rates across all test instances from all tasks. \method{}-VT and \method{} show significant improvement in performance over the vanilla ViT.}
\label{fig:arc-task-solve-distribution}
\end{figure}

In this paper, we explore the potential of vision transformers to solve ARC tasks using supervised learning. 
We assess how well transformers can learn complex mappings for a single task when provided with sufficient training data. Our exploration highlights fundamental representational limitations of vision transformers on the ARC, leading to three high-level findings that we believe provide a strong foundation for future research in visual reasoning using transformer-based architectures:
\begin{enumerate}[leftmargin=*]
    \item \textbf{A vanilla Vision Transformer (ViT) fails on the ARC:} Despite the ARC grids' relatively simple structure compared to the much larger, noisier natural images they are typically evaluated on, a vanilla ViT performs extremely poorly on $90\%$ of the tasks with an overall test accuracy of $18\%$~(cf. \Cref{fig:arc-task-solve-distribution},~\Cref{sec:defaultVit}). This is despite using a training set of one million examples per task. Following a failure analysis, we hypothesize that the vanilla ViT fails because it cannot accurately model spatial relationships between the objects in an ARC grid and the grid boundaries. 
    \item \textbf{A 2D visual representation significantly boosts ViT reasoning performance:} Using a 2D representation strategy based on \textit{visual tokens} to represent the ARC input-output pairs, \method{} solves $66\%$ of all test instances -- a marked improvement~(cf. Section~\ref{sec:VisualTokens}). About 10\% of the tasks remain poorly solved. After further failure analysis on these tasks, we discover that certain complex visual structures are 
    difficult for \method{}.  We hypothesize this is due to limitations of the transformer architecture itself in that it  is designed to prioritize token embeddings over positional encodings that can make it challenging to capture intricate spatial relationships.
    \item \textbf{Positional Information further enhances ViT reasoning abilities:} We improved~\method{}'s spatial awareness by learning to combine absolute, relative, and~\textit{object} positional information~(cf. Section~\ref{sec:PositionalEnhancements}), resulting in substantial performance boosts, with some ARC tasks progressing from unsolved to fully solved (\Cref{fig:arc-task-solve-distribution}). The final test accuracy is $75\%$, with more than half of the tasks being solved to an accuracy of $95\%$ or more.
\end{enumerate}

\footnotetext[2]{\label{fn:taskA}Task A follows a rule based on color count: if the input grid has two distinct colors, the output contains a grey diagonal from the top-left to the bottom-right. Conversely, if the input grid has three colors, the grey diagonal is from the top-right to the bottom-left.}

\section{Related Work}


\paragraph{Abstract Visual Reasoning (AVR)} \!\!\!\!\! is an emerging field that seeks to measure machine ``intelligence''~\citep{malkinski2023review}. Unlike many popular studies that focus on visual reasoning with multi-modal input~\citep{vqa,clevr,vcr,pr,vqai}, AVR focuses on reasoning tasks where the inputs are strictly images. The goal of AVR tasks is to discover abstract visual concepts and apply them to new settings. While the ARC is a generation task using abstract rules, other AVR tasks include classification tasks with explicit rules, such as the Raven’s Progressive Matrices~\citep{raven} and Odd-One-Out~\citep{ooo}. We refer the readers to \citet{malkinski2023review} for a more detailed introduction to AVR.

\paragraph{Vision Transformers \& Positional Encoding.}
\label{sec:ViTandPE}

A Transformer architecture is based on the
attention mechanism~\citep{attention-all-u-need}. Following successes in natural language processing~\citep{gpt3,gpt4,bert}, recent studies have extended the Transformer to the vision domain~\citep{vit-survey}. State-of-the-art approaches involve dividing the image into rectangular ``patches''\citep{vit}, where various techniques such as dynamic patch sizes allow for more effective capture of local information~\citep{msvit, dynamic-vit}. Vision Transformers have been successfully used to perform various image-to-image generation tasks such as inpainting \citep{transformer-inpaint},  image restoration \citep{transformer-super-resolution}, colorization \citep{transformer-colorization}, and denoising \citep{transformer-denoise}.

Due to the set-based (permutation-invariant) nature of attention, Positional Encodings are used to inject positional information in a Transformer~\citep{attention-all-u-need}. State-of-the-art Positional Encodings include Absolute Positional Encodings (APEs) where unique encodings are added to the inputs directly~\citep{bert},  Additive Relative Positional Encodings (RPEs)~\citep{rpe, t5, fire-rpe} that measure the relative positions between tokens by modifying the attention logits, and various hybrid methods~\citep{rope,fire+random}. Vision Transformer research has adapted these concepts, implementing both APEs~\citep{vit} and RPEs~\citep{2drpe-vit} to incorporate positional information about the image patches.

\paragraph{Solvers for the ARC.}




Since the introduction of the ARC~\citep{chollet2019measure}, the development of solvers has been an active research area. The earliest successful approaches consisted of an expressive Domain Specific Language (DSL) and a program synthesis algorithm that searched for a valid solution program expressed in the DSL. These include DAG-based search~\citep{arcsolver-kaggle}, graph-based constraint-guided search~\citep{arga}, grammatical evolution~\citep{arcsolver-grammar-evo}, library learning~\citep{arcsolver-dreamcoder}, compositional imagination~\citep{arcsolver-imagination}, inductive logic programming~\citep{arcsolver-ilp}, decision transformers~\citep{arcsolver-decision-transformer}, generalized planning~\citep{arcsolver-planning}, reinforcement learning~\citep{arcsolver-arcle}, and several others~\citep{arcsolver-vips, arcsolver-descriptive-grid}. These models achieved up to 30\% on the private ARC test set~\citep{kaggle-comp, arcathon}.

A few ARC solvers do incorporate CNNs for limited subtasks, e.g.\ recognizing background colors~\citep{arcsolver-llm2} 
or extracting simple image features~\citep{bober2024neural}, but none attempt the broader visual transformations 
that ARC demands. In contrast, recent efforts have centered on Transformer-based Large Language Models (LLMs), 
which were shown to exhibit an apparent ability to perform “reasoning”~\citep{cot}. Such methods were prompted to perform program synthesis on a DSL~\citep{arcsolver-llm-syn, arcsolver-hysynth} as well as general-purpose languages such as Python~\citep{arcsolver-codeit, arcsolver-hyp-search}, with the best-performing model achieving 42\% on the public ARC evaluation set~\citep{arcsolver-greenblatt}. LLMs were also explored as standalone solvers, where they were asked to produce the output grids directly instead of outputting a program. Although pre-trained LLMs proved ineffective when generating the output grid pixels directly~\citep{arcsolver-llm2, arcsolver-deepmind, conceptarc}, its performance was shown to be improved by object representation~\citep{arcsolver-llm} and test-time fine-tuning~\citep{arcsolversota, akyurek2024surprising}. The vision variant of a state-of-the-art LLM, GPT-4V was shown to be ineffective~\citep{gpt4v, arcsolver-llm}.


OpenAI’s o3 system reportedly reached 87.5\% on a semi-private ARC evaluation~\citep{o3arc},
though its model architecture and computational budget remain undisclosed, and it was not evaluated on the official private test set. At the time of writing, the solver publicly recognized as state-of-the-art was the winner of the ARC Prize 2024 challenge~\citep{chollet2024arc} achieving 53.5\% on the private test set using a publicly available \emph{pipeline-based} approach~\citep{da-fr-arc2024}
This pipeline fine-tunes a Mistral-NeMo-Minitron-8B-Base model, applies data augmentations, and performs an additional round of test-time training and heuristic candidate selection.



This work complements recent advances in LLM-based ARC solvers, which primarily rely on 1D token representations~\citep{arcsolver-llm} and thus underutilize the inherently two-dimensional nature of ARC tasks. In contrast, we propose an orthogonal, vision-centric strategy grounded in a specialized Vision Transformer that explicitly captures spatial structure. By addressing a fundamental shortcoming of existing 1D pipelines, our method can seamlessly integrate with LLM-based solutions and extend to a broader range of visual reasoning tasks that require precise 2D modeling.



\section{Vanilla Vision Transformer for the ARC: An Initial Approach}
\label{sec:defaultVit}

We first implement a vanilla Vision Transformer architecture as detailed in~\citet{vit} and~\citet{deit} as a solver for the ARC. 
Consider an input image \(I\) divided into \(P \times P\) non-overlapping patches. Each patch \(p_i\) is flattened in raster order and indexed by \(i\) before being projected into a \(d\)-dimensional embedding space. Let \(h_i^0\) denote the initial input to the Transformer for patch \(p_i\). For the $n$-th Transformer layer, \(n \in\{ 1, \dots, N\}\), and for a single attention head, the following operations are performed:
\begin{align}
    h_{i}^{0} &= \mathbf{E}_{p_{i}} + \mathbf{E}_{\text{pos}_{i}} \label{eq:input_embedding} \\
    \hat{h}_{i}^{n} &= \text{LayerNorm}(h_{i}^{n-1})\\
    q_{i}^{n}, k_{i}^{n}, v_{i}^{n} &= \hat{h}_{i}^{n} \mW_{q}^{n}, \quad 
    \hat{h}_{i}^{n} \mW_{k}^{n}, \quad 
    \hat{h}_{i}^{n} \mW_{v}^{n} \label{eq:qkv_projections} \\
    A_{i,j}^{n} &= \frac{q_{i}^{n}\cdot k_{j}^{n}}{\sqrt{d}}  \label{eq:attention_score} \\
    o_{i}^{n} &= \sum_{j} \text{Softmax}(A_{i,j}^{n}) v_{j}^{n} + h_{i}^{n-1} \label{eq:attention_output} \\
    f_{i}^{n} &= \text{FeedForward}(\text{LayerNorm}(o_{i}^{n})) + o_{i}^{n} \label{eq:feedforward} \\
    h_{i}^{n} &= \text{LayerNorm}(f_{i}^{n}) \label{eq:residual_connection}
\end{align}

Here, \( \mathbf{E}_{p_{i}} \) is the embedding of patch \( p_i \) and \( \mathbf{E}_{\text{pos}_{i}} \) is the positional encoding. Following the standard ViT implementation of~\cite{vit}, the Absolute Positional Encoding (APE) is calculated as a learnable 1D encoding:
\[
\mathbf{E}_{\text{pos}_i} = \mW_i, \quad \mathbf{E}_{\text{pos}_i} \in \mathbb{R}^{d}, \quad \mW \in \mathbb{R}^{L\times d}
\]
where \( \mW \) is a learned matrix assigning a $d$-dimensional vector to each of the possible \( L \) positions; $L$ is the maximum input length.

As seen in~\Cref{fig:arc-examples}, ARC tasks are~\textit{generative} and require mapping an input image to an output image. Because image dimensions may vary across instances of the same task and even between the input and output grids of the same instance, any model that generates candidate solutions to an ARC input must be able to ``reason'' at the pixel level. We adapt the ViT architecture to this setting by making the following key modifications:
\begin{itemize}[leftmargin=*]
    \item[--] We introduce a decoder with cross-attention using the same positional encoding and attention mechanisms of the encoder. After the final decoder layer \( N \), the output embedding \( h_{i}^{N} \) of patch $i$ is  projected linearly and a softmax function is applied to predict pixel-wise values \( \hat{y}_i \) as
    $\hat{y}_i = \text{Softmax}(\text{Linear}(h_{i}^{N}))$.
The cross-entropy loss is computed as the sum over pixels, $-\sum_i y_i \log(\hat{y}_i)$.

    \item[--] To achieve the required pixel-level precision for the ARC task, we employ a patch size of \( 1 \times 1 \), effectively treating each pixel as an independent input token.
    \item[--] To handle variable-sized grids, the flattened list of tokens is padded to a fixed maximum length. This configuration enables the model to process and generate ARC task outputs pixel-by-pixel.
\end{itemize}

\subsection{Experiments}
\paragraph{Data.} To evaluate ViT's reasoning capabilities comprehensively, we treat each of the 400 public training ARC tasks as an individual AVR problem. We generate a dataset of 1 million input-output pairs per task using the RE-ARC generator~\citep{hodel_rearc_2024} and train all of our models (the vanilla ViT and~\method{} models) in a supervised manner from scratch. For experiments involving reduced training sets, please see 
Appendix~\ref{appendix:data_ablation} for performance under varying data-regime settings.


\paragraph{Hyperparameters and training protocol.} The ViT baseline consists of three layers with eight attention heads and a hidden dimension of 128. We trained the model on various single-core GPU nodes, including P100, V100, and T4, using a batch size of 8 for one epoch. We chose to train for one epoch because most models showed signs of convergence within the epoch. Due to computational resource limitations, we evaluated our major milestone models on the full set of 400 tasks. However, for the ablation studies hereafter, we used a randomly sampled subset of 100 tasks. For more details on the training process, please refer to~\Cref{appendix:training_detail}. Our code is available in the supplementary materials and will be open-sourced upon publication.

\paragraph{Evaluation metric.} We evaluate the model primarily on the percentage of solved instances, using a strict criterion: an instance is considered solved only if all generated pixels, including padding and border tokens, exactly match the ground truth. This approach is stricter than the original ARC metric which permits up to three candidate solutions.

\paragraph{Results.}
\Cref{fig:arc-task-solve-distribution} shows that the vanilla ViT performs poorly: a significant percentage of tasks have a near 0\% solve rate despite the million training examples per task. This points to fundamental limitations of the ViT architecture that inhibit abstract visual reasoning. In the following sections, we analyze failure cases and investigate methods for enhancing the visual reasoning ability of ViT.

\section{Visual Tokens: a Better Representation for ViT}
\label{sec:VisualTokens}


The basic version of our~\method{} framework builds on the vanilla ViT but includes three simple yet highly effective changes to the representation of the ARC grids. We refer to these changes as \textit{visual tokens} to emphasize a departure from the language-based tokenization perspective in the particular setting of the ARC.

\begin{figure}
    \centering
    \includegraphics[width=0.6\textwidth]{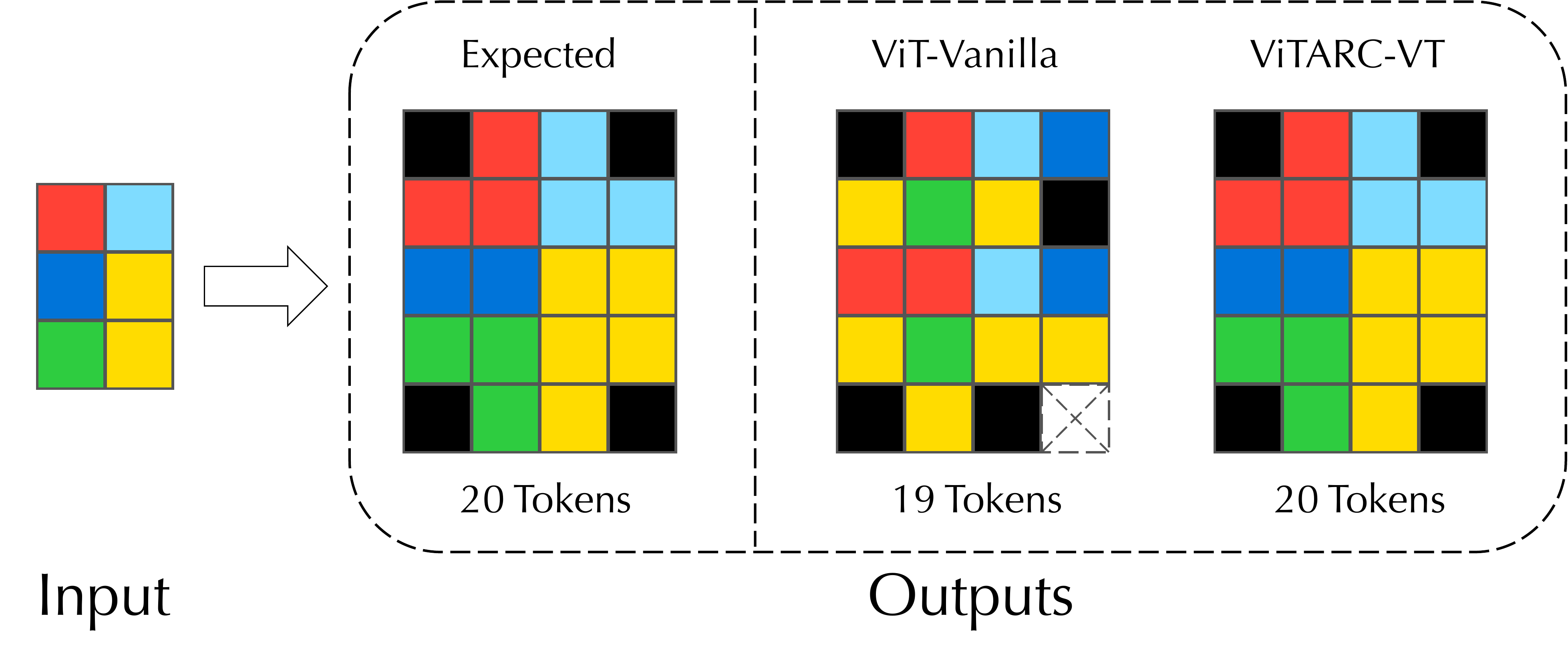}
    \caption{\textbf{Visualization of ViT-Vanilla failure case (Task B from \cref{fig:arc-examples}).} The output of ViT-Vanilla has an incorrect number of tokens (19) compared to the expected 20. For better visualization, the output pixels are arranged to match the grid, although the model generates the pixels in a continuous sequence in raster order. This makes the task a relaxed output prediction for ViT-Vanilla, where the flattened output sequence is compared with the expected output sequence.}

    \label{fig:vit-failure}
\end{figure}



\paragraph{2D padding.} 
We observed that a large portion of the incorrect outputs from the vanilla ViT had incorrect grid sizes, a flagrant failure mode; An example is visualized in \Cref{fig:vit-failure} (ViT-Vanilla). We hypothesize that this is due to the vanilla ViT implementing padding in a ``1D'' manner, where \texttt{<pad>} tokens are applied to the sequence after flattening, thus losing the two-dimensional context. To address this issue, we implemented 2D padding, where \texttt{<pad>} tokens are applied to the image \textit{first} before being flattened in raster order into a sequence for transformer processing (see~\Cref{fig:overview}).

However, this design introduces a new drawback: the model must now predict \texttt{<pad>} tokens as part of the output grid. In initial experiments, we observed that the model tends to ignore these \texttt{<pad>} tokens (that do not receive attention), erroneously predicting over the entire \( h_\text{max} \times w_\text{max} \) grid rather than focusing on the valid input region. An example of this issue is shown in ~\Cref{fig:defaulViT_failure} of~\Cref{appendix:default-vit-fail}. To address this, we define \texttt{<2d\_pad>} tokens and enable attention to these tokens, allowing the model to properly account for the padded regions 
as well as the valid output region.

\paragraph{Border tokens for spatial awareness.} The implementation of 2D padding did not completely alleviate the previously observed failure cases. We further observed that for some tasks, when the output is cropped to the true grid dimensions, the predictions within the valid region are correct, underscoring the importance of proper boundary handling. We show an example in~\Cref{fig:defaulViT_failure} of~\Cref{appendix:default-vit-fail}. Inspired by the use of end-of-sequence (EOS) tokens like \texttt{</s>} in Natural Language Processing (NLP), we introduce \textit{border tokens} to explicitly define the grid boundaries (cf.~\Cref{fig:overview}):

\begin{itemize}
   \item[--] \textbf{Newline tokens} (\texttt{\footnotesize <2d\_nl>}) mark row transitions in the \(h_\text{max} \times w_\text{max}\) grid.
   \item[--] \textbf{End-of-grid tokens} (\texttt{\footnotesize <2d\_endxgrid>}, \texttt{\footnotesize <2d\_endygrid>}, and \texttt{\footnotesize <2d\_endxygrid>}) delineate the true \(h \times w\) grid boundaries.
\end{itemize}

The introduction of border tokens enables the model to more effectively distinguish the task grid from the padding. Without these tokens, the model would need to count tokens to determine boundaries, which becomes unreliable—especially in ARC tasks with dynamically defined output grid sizes (e.g., task C in~\Cref{fig:arc-examples}). Furthermore, as we see in ViT-Vanilla failure cases (\Cref{fig:vit-failure}), it is ambiguous to recover the 2D positions from a 1D sequence of predicted tokens alone. Border tokens also provide a fixed 2D template to fill in, which implicitly helps reconstruct the correct 2D positions and makes it easier to debug the related grid logic.




\paragraph{2D Absolute Positional Encoding.} With the introduction of 2D padding and border tokens, our setup now operates on fixed-size, two-dimensional input-output pairs that are aligned with a universal \((x, y)\) coordinate system. This allows us to adopt existing positional encoding (PE) strategies from the literature (see Section~\ref{sec:ViTandPE}). After empirical analysis, we implement a (non-learned) 2D sinusoidal APE for \method{}, which is defined as follows:
\begin{equation}
\text{Sinusoid}(p) = 
\begin{bmatrix}
\sin\left(\frac{p}{10000^{2k/d}}\right) \\
\cos\left(\frac{p}{10000^{2k/d}}\right)
\end{bmatrix}, \quad k=0,\ldots, d/2,
\label{eq:sinusoid}
\end{equation}
\begin{equation}
\mathbf{E}_{\text{pos}_{(x, y)}} = \text{concat}\left(\text{sinusoid}(x), \text{sinusoid}(y)\right),
\label{eq:2d-ape}
\end{equation}

where $p$ represents either the $x$ or $y$ coordinate, $k$ is the index of the positional encoding dimension, and $d$ is the total embedding dimension.

\subsection{Results}

\begin{figure}[t]
\begin{center}
\includegraphics[width=0.8\linewidth]{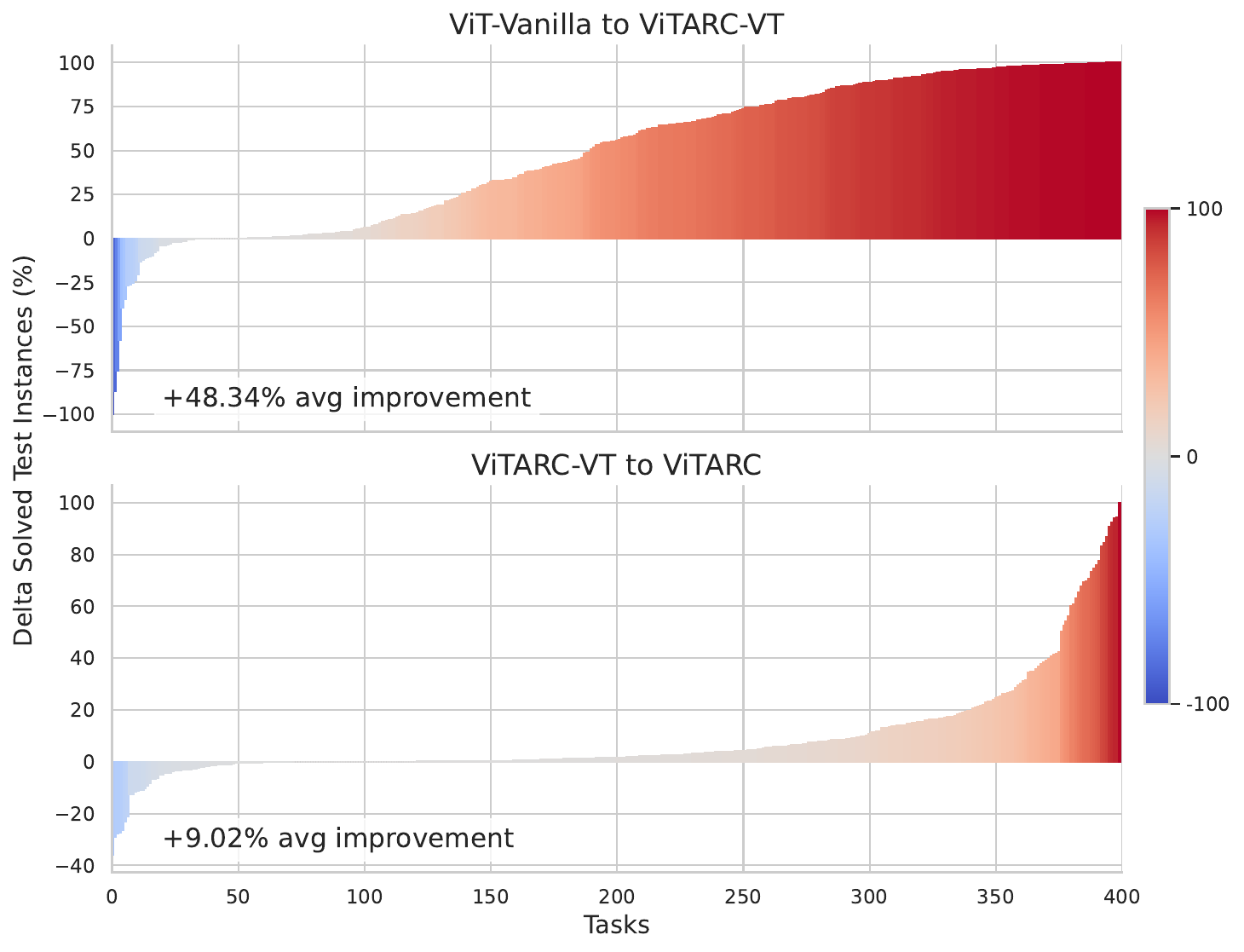}
\end{center}
\caption{\textbf{Improvement in percentage of solved test instances per task.} (a) From ViT-Vanilla to ViTARC-VT: We observe that over 85\% of tasks benefit from the introduction of 2D Visual Tokens, showing consistent gains compared to the vanilla ViT. (b) From ViTARC-VT to ViTARC: We observe that more than half of all tasks show further improvement. Improvement from ViT-Vanilla to ViTARC is shown in \Cref{fig:pertask_full} in \Cref{appendix:full_results} where a 57.36\% average improvement is observed.}

\label{fig:inc_ablation_pertask}
\end{figure}

\Cref{fig:arc-task-solve-distribution} shows substantial improvements in test accuracy due to the 2D visual tokens just described.~\Cref{fig:inc_ablation_pertask}(a) illustrates the improvement in the percentage of solved instances for each task. We observe an average performance boost of 48.34\% compared to the baseline ViT across the 400 tasks. This model, referred to as ViTARC-VT, demonstrates that the new representation with 2D visual tokens significantly enhances the model’s ability to handle AVR tasks. 

A key driver of this improvement is the use of 2D padding, which creates a fixed schema for 2D positions. This ensures consistent spatial alignment and effectively addresses the challenge of applying 2DAPE to variable-sized grids, where unknown output positions during inference complicate accurate mapping.


To quantify the contribution of border tokens, we performed an ablation study. As seen in~\Cref{fig:substep_ablation}, the absence of border tokens leads to a 4.59\% decrease in accuracy, emphasizing their importance in helping the model delineate task grid boundaries and maintain spatial consistency in the input representation. For more detailed numerical results, refer to \Cref{tab:sample100_table} in \Cref{appendix:substep_ablation}.

\subsection{Analysis}

\begin{figure}[t]
\begin{center}
\includegraphics[width=1.0\linewidth]{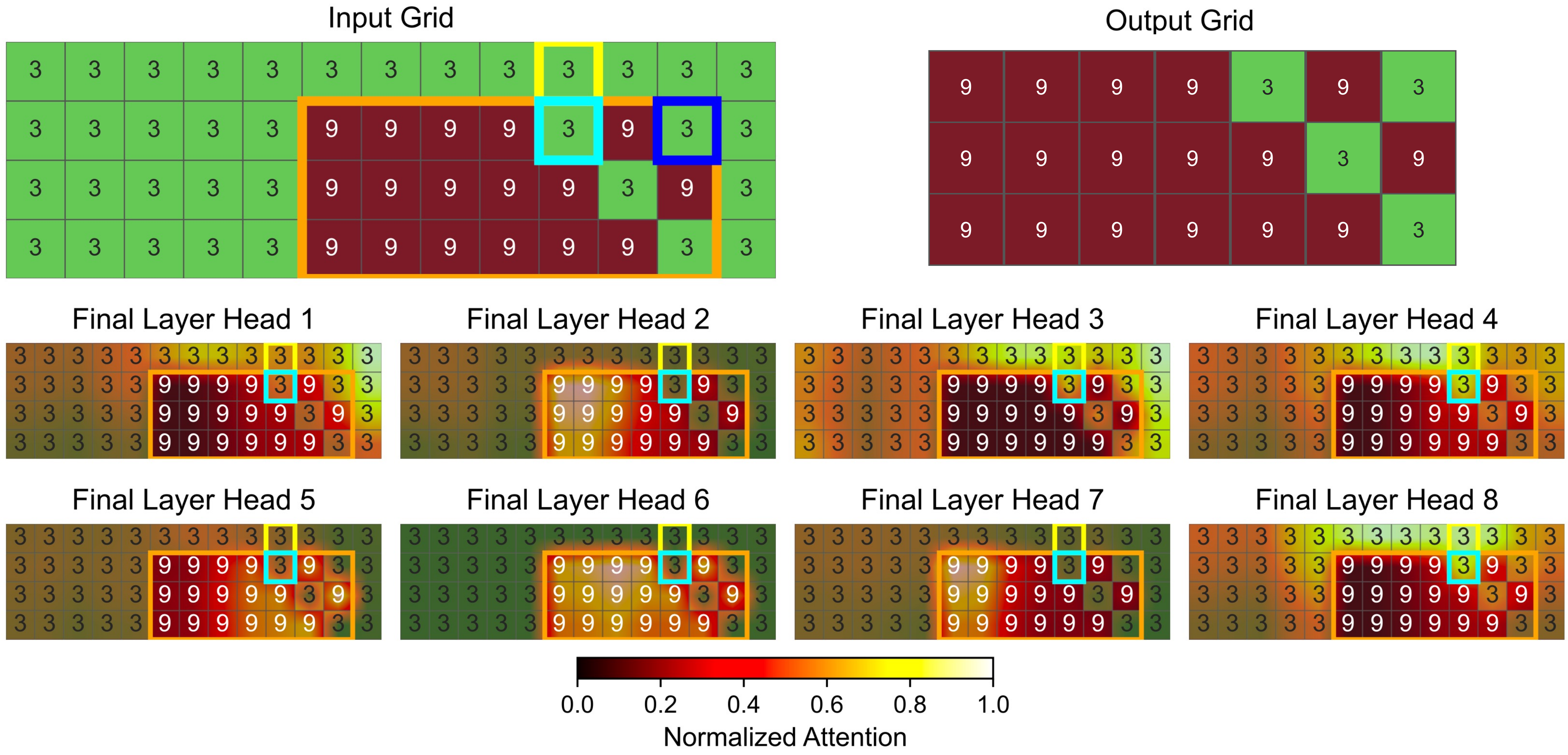}
\end{center}
\caption{\textbf{\method{}-VT failure analysis for ARC task (\#1cf80156):} Cross-attention thermal heatmaps from all attention heads in the final layer at the step predicting the color-3 pixel (dark blue box). The task demands identifying the largest rectangular subgrid (orange bounding box), yet the visualized attention shows that no heads effectively distinguish this subgrid from its surroundings, indicating the need for stronger positional cues (\textit{PEMixer}). Moreover, the color-3 pixel inside the cyan box (within the subgrid) is not easily differentiated from the pixel in the yellow box (outside the subgrid) by 2DAPE, underscoring the motivation for \textit{2D-RPE} and \textit{OPE} to provide more precise spatial biases.}
\label{fig:cross_attn}
\end{figure}

While ViTARC-VT delivers strong results—approximately 40\% of ARC tasks achieved over 90\% solved test instances—there remain certain tasks where the model struggles. Specifically, around 10\% of ARC tasks have less than 5\% of test instances solved, even after training on a large dataset containing one million examples per task. Closer examination reveals that tasks involving complex visual structures, such as concave shapes, holes, or subgrids, are consistently problematic. These challenges highlight certain architectural limitations, particularly the model's difficulty in segmenting multi-colored objects, where positional information should ideally play a more dominant role.

To better understand this behavior, we refer back to~\Cref{eq:input_embedding}: $h_{i}^{0} = \mathbf{E}_{{p}_{i}} + \mathbf{E}_{\text{pos}_{i}}$. 
In this setup, the absolute positional encoding, $\mathbf{E}_{\text{pos}_{i}}$, is directly added to the input embedding, $\mathbf{E}_{{p}_{i}}$, so that it adjusts the token's representation without overwhelming its semantic content. This works effectively in NLP tasks, where the semantic meaning of tokens generally takes precedence over their position. However, in vision tasks, especially those requiring detailed visual reasoning, spatial relationships often carry as much importance as, if not more than, the content of the tokens. For tasks in the ARC that involve complex multi-colored objects, such as subgrids, accurately encoding positional information becomes crucial.
\Cref{fig:cross_attn} illustrates a specific case where the model fails to group pixels within a multi-colored subgrid correctly. The cross-attention map reveals that the model overly relies on color similarity, resulting in confusion between similarly colored pixels in different positions. This indicates a lack of sufficient attention to spatial relationships, which is essential for such tasks and guides us to develop further enhancements in the next section.



\section{Recentering Positions \& Objects for Spatial Reasoning in ViT}
\label{sec:PositionalEnhancements}

Our observations on the failure cases of ViTARC-VT lead us to implement further enhancements to tackle tasks with complex visual structures by better encapsulating the positional information of pixels and objects.  

\paragraph{Positional Encoding Mixer (PEmixer).}
To better balance the importance of positional information and tokens, we modify~\Cref{eq:input_embedding} by learning weight vectors for the encodings, i.e.,
\begin{equation}
h_{i}^{0} = \boldsymbol{\alpha} \odot \mathbf{E}_{{p}_{i}} + \boldsymbol{\beta} \odot \mathbf{E}_{\text{pos}_{i}},
\label{eq:mixer}
\end{equation}

where $\boldsymbol{\alpha}$ and $\boldsymbol{\beta}$ are \textbf{learnable} vectors of the same dimension as the encoding vectors, and $\odot$ denotes element-wise multiplication. This effectively allows the model to learn the optimal balance between input tokens and positional encoding.


Furthermore, our implementation of 2D APE as described in~\Cref{sec:VisualTokens}, where $\mathbf{E}_{\text{pos}_{(x,y)}}$ is the concatenation of $\mathbf{E}_{\text{pos}_{x}}$ and $\mathbf{E}_{\text{pos}_{y}}$, allows the vector-based mixing coefficients to focus on specific coordinates, which further improves the model's reasoning capability over specific pixels.
For details on our PEmixer strategies and empirical evidence that PEmixer boosts positional information, 
please refer to Appendix~\ref{appendix:pemixer_strategies}.

\paragraph{2D Relative Positional Encoding (2D-RPE).}

Motivated by the example in~\Cref{fig:cross_attn}, we aim to enable the model to distinguish between pixels in different spatial regions, such as the color-3 (green) pixel in the cyan box versus the one in the yellow box. In this example, the positional difference between the two pixels is just 1 along the \(y\)-coordinate. APE encodes this difference as a small shift; while the transformer is theoretically capable of capturing these spatial relationships, in practice often requires many training epochs~\citep{hahn2020theoretical}.

To better account for spatial relationships in two-dimensional grids, we adapt the Relative Positional Encoding (RPE) approach from ALiBi~\citep{alibi-rpe} and extend it to 2D. ALiBi introduces additive positional biases to the attention scores based on the relative positions of tokens. In its original 1D form, ALiBi defines the positional bias as the following:
\begin{equation}
\begin{aligned}
A_{i,j}^{n} = \frac{q_{i}^{n}\cdot k_{j}^{n}}{\sqrt{d}} + \mathbf{B}_{\mathbf{P}_{i,j}}, \quad
\mathbf{B}_{\mathbf{P}_{i,j}} = r \cdot |i - j|,
\end{aligned}
\label{eq:1d_alibi_rpe}
\end{equation}
where \(\mathbf{P}_{i,j}\) represents the relative positional offset between tokens \(i\) and \(j\), and \(r\) is a predefined slope that penalizes tokens based on their distance.

Extending to 2D, we introduce distinct slopes for the ``left'' and ``right'' directions, efficiently capturing directional biases along the x and y axes. This design leverages the inherent 2D structure of the data while aligning with the sequential raster order of the generation process. Specifically:
\begin{itemize}
    \item[--] Pixels located above or to the left of the current pixel in 2D space are assigned a bias $r_{\text{before}}$. 
    \item[--] Pixels located below or to the right are assigned a bias $r_{\text{after}}$. 
\end{itemize}

Hence, the 2D-RPE bias is computed as:
\begin{equation}
  \mathbf{B}_{\mathbf{P}_{i,j}} = 
  \begin{cases} 
    r_{\text{before}} \cdot d\left((x_i, y_i), (x_j, y_j)\right), & \text{if } j \leq i, \\
    r_{\text{after}} \cdot d\left((x_i, y_i), (x_j, y_j)\right), & \text{if } j > i,
  \end{cases}
\end{equation}
where \( d\left((x_i, y_i), (x_j, y_j)\right) \) represents the 2D Manhattan distance between coordinates $(x_i, y_i)$ and $(x_j, y_j)$. The slope values \( r_{\text{before}} \) and \( r_{\text{after}} \) are derived following the ALiBi setup, forming a geometric sequence of the form \( 2^{-8/n} \) for \( n \) heads. \( r_{\text{before}} \) starts at \( 1/2^1 \), while \( r_{\text{after}} \) starts at \( 1/2^{0.5} \), both using the same ratio.

In this work, we leverage both 2D-RPE and 2D sinusoidal APE within our model. In contrast to observations made in hierarchical ViTs like Swin~\citep{swin}, where a degradation in performance was noted when combining RPE with APE, our results demonstrate a marked improvement. The inclusion of 2D-RPE allows for more precise modeling of relative spatial relationships, complementing the global positional information provided by APE. This synergy proves particularly effective for tasks demanding fine-grained spatial reasoning. 

Further details on alternative 2D RPE configurations, as well as experimental comparisons on a subset of 50 ARC tasks, can be found in the Appendix~\ref{appendix:appendix_rpe}.


\paragraph{Object-based Positional Encoding (OPE).}
\label{sec:OPE}

For tasks involving multi-colored objects, or more generally, tasks that require objectness priors \citep{chollet2019measure}, external sources of knowledge about object abstractions can be integrated into the model. We inject this information through a novel \emph{object-based positional encoding}.
We extend the 2D sinusoidal APE defined in~\Cref{eq:2d-ape} by introducing the object index $o$ as an additional component to the pixel coordinates $(x, y)$. This results in a modified positional encoding:
\begin{equation}
\mathbf{E}_{\text{pos}_{(o, x, y)}} = \text{concat}\left(\text{sinusoid}(o), \text{sinusoid}(x), \text{sinusoid}(y)\right).
\end{equation}
In object detection models, two primary segmentation methods are bounding box segmentation and instance segmentation, the latter of which captures precise object boundaries. For simplicity, we adopt bounding box segmentation to derive the object index $o$, as fine-grained distinctions at the instance level can already be addressed by the model’s attention mechanism, as illustrated in~\Cref{fig:cross_attn}. \Cref{fig:overview} demonstrates how bounding box information is obtained and incorporated into the positional encoding.

This design integrates seamlessly with the PEmixer introduced earlier, as it enables the model to dynamically adjust its reliance on the object index $o$ based on the task's needs. In scenarios where the object index provides valuable abstraction, the model can prioritize it, while in cases where the object-based method is less effective, the model can fall back on the $(x, y)$ positional information.

For our experiments, OpenCV’s contour detection~\citep{opencv_library} proved sufficient for generating object indices in the ARC tasks, demonstrating the practical effectiveness of OPE. This novel approach not only addresses challenges related to complex object shapes but also establishes a method for injecting external objectness knowledge into vision models, enhancing their reasoning capabilities. Moreover, more advanced or domain-agnostic segmentors—such as SAM—could be integrated seamlessly to handle larger grid cells or complex, real-world settings, thereby further extending the applicability of our approach.

\subsection{Results}

\begin{figure}
    \centering
    \includegraphics[width=0.85\linewidth]{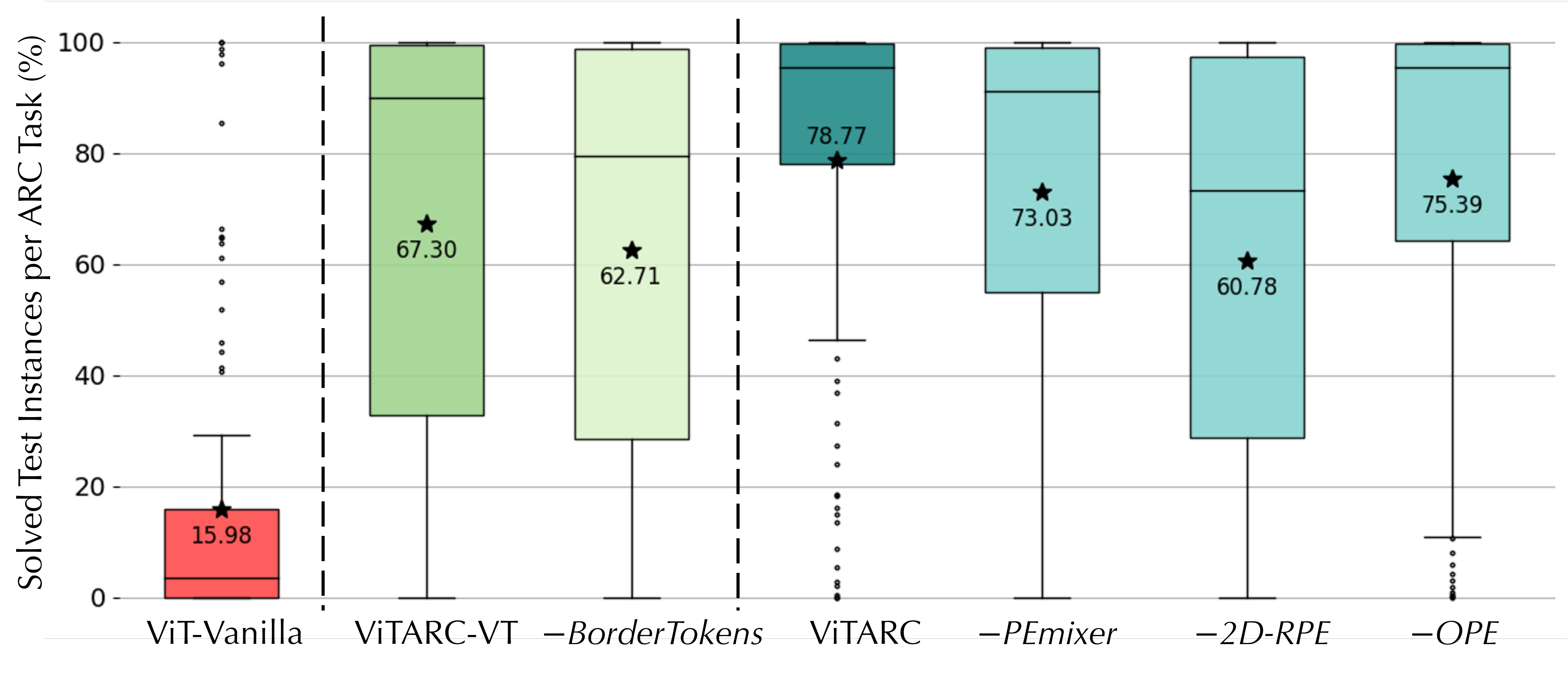}
    \caption{\textbf{Distribution statistics of solve rates on 100 random tasks for ablation.} 7 Models are shown: ViT-Vanilla, ViTARC-VT, and ViTARC are the models introduced in Sections \ref{sec:defaultVit}, \ref{sec:VisualTokens} and \ref{sec:PositionalEnhancements} respectively.  Ablated components are prefixed as $-$ and ablate the full model to the left, i.e., $-\textit{BorderTokens}$ is an ablation of this component from ViTARC-VT and each of $-\textit{PEmixer}$, $-\textit{2D-RPE}$, and $-\textit{OPE}$ ablate these respective components from ViTARC.} 
    \label{fig:substep_ablation}
\end{figure}




We arrive at our final model, ViTARC, which contains all the improvements mentioned in \Cref{sec:VisualTokens} and \Cref{sec:PositionalEnhancements}. The final encoding combines all three components: 2DAPE, 2DRPE, and OPE, leveraging their complementary strengths to enhance spatial reasoning. As shown in \Cref{fig:arc-task-solve-distribution}, the model is a significant improvement over both the baseline ViT-Vanilla and ViTARC-VT due to the proposed positional enhancements.

Furthermore, \Cref{fig:inc_ablation_pertask}(b) highlights the generalization of these improvements across tasks, with an additional 9.02\% increase in solved instances compared to ViTARC-VT. ViTARC-VT itself already achieved a significant boost over ViT-Vanilla, culminating in a total improvement of 57.36\% over the baseline ViT-Vanilla.

\Cref{fig:substep_ablation} further illustrates the impact of each enhancement on task performance. All three contribute to the overall improvement, with 2D-RPE providing the largest gain, followed by PEmixer and OPE. Notably, without 2D-RPE, the model’s performance drops below that of ViTARC-VT. This occurs because OPE, while effective in specific tasks, is not consistently reliable. In these cases, ViTARC must fall back on the $(x, y)$ embeddings from 2D-APE, which are less expressive due to their lower dimensionality compared to ViTARC-VT. The inclusion of 2D-RPE recovers these positional signals at the attention level, ensuring robust performance even when object-based cues are insufficient.

For a comprehensive breakdown of the task-level performance and the numerical details of these ablations, please refer to \Cref{appendix:substep_ablation} and \Cref{appendix:ablations_pemixer_2drpe_ope}.

\section{Discussion \& Limitations}
\label{sec:discussion}

\textbf{On CNNs.}
While CNNs excel at capturing local pixel patterns, they lack explicit positional encodings and can
struggle with the object-centric transformations required by ARC. Nonetheless, integrating CNNs as
tokenizers or feature extractors before an attention mechanism remains an intriguing avenue, potentially
combining local convolutions with the global reasoning that Transformers provide.

\textbf{Beyond ARC.}
Our work focuses on ARC as a controlled environment to diagnose fundamental ViT shortcomings
in abstract visual reasoning. We believe, however, that the principles outlined—explicit 2D positional
modeling, object-based positional encodings, and attention-based spatial transformations—could transfer
to other AVR benchmarks, such as Raven’s Progressive Matrices (RPM), provided one adapts patch
sizes. We leave these extensions
to future work.


\section{Conclusion}

This paper introduced \method{}, a Vision Transformer architecture designed to address the unique challenges posed by the Abstraction and Reasoning Corpus. A key finding of our work is that positional information plays a critical role  in visual reasoning tasks. While often overlooked when adapting transformers from NLP to vision, our results demonstrate that even simple enhancements to positional encoding can significantly improve performance on ARC tasks. Furthermore, we show that incorporating object indices as additional positional information via OPEs provides a meaningful improvement in handling complex spatial relationships in ARC tasks.

Additionally, we introduced 2D padding and border tokens to handle variable-sized images requiring high precision in visual reasoning. Given ARC’s pixel-level precision and abstract reasoning requirements (e.g., $1 \times 1$ pixel tasks in ARC, but potentially $n \times n$ pixels in more generalized visual reasoning), resizing or cropping—commonly used in standard vision tasks—is infeasible. \method{} reveals limitations in current ViT structures under these conditions and suggests necessary adaptations for such tasks.


It is important to note that \method{} solves task-specific instances of ARC in a data-driven approach, treating each ARC task independently. This method does not fully solve ARC, which requires the ability to generalize across different tasks—a challenge that remains open for future research. However, since the current state-of-the-art (SOTA) in ARC relies on LLM-based transduction models that handle tasks through supervised input-output transformations~\citep{chollet2024arc}, integrating the 2D inductive bias from ViTARC could provide an orthogonal benefit. This is especially relevant as prior studies indicate that the sequential nature of 1D methods in LLMs can limit ARC performance; for example, because the input grid is processed in raster order, LLMs experience a significant drop in success rates when horizontal movement/filling tasks are rotated 90 degrees~\citep{arcsolver-llm}.

In summary, this work highlights the importance of 2D positional information and object-based encodings in abstract visual reasoning that leads to our novel contribution of the \method{} architecture. \method{} advances the application of Vision Transformers for pixel-level reasoning and suggests further avenues for improving generalization capabilities in models tackling 
visual reasoning tasks.




\subsubsection*{Acknowledgments}

This work was supported by the Institute of Information \& Communications Technology Planning \& Evaluation (IITP) grant funded by the Korean Government (MSIT) (No. RS-2024-00457882, National AI Research Lab Project).
Elias B.~Khalil acknowledges support from the SCALE AI Research Chair program.

\clearpage
\bibliography{main}
\bibliographystyle{tmlr}

\clearpage
\appendix

\section{Vanilla ViT Failure Analysis}
\label{appendix:default-vit-fail}

\begin{figure}[htbp!]
\begin{center}
\includegraphics[width=0.6\linewidth]{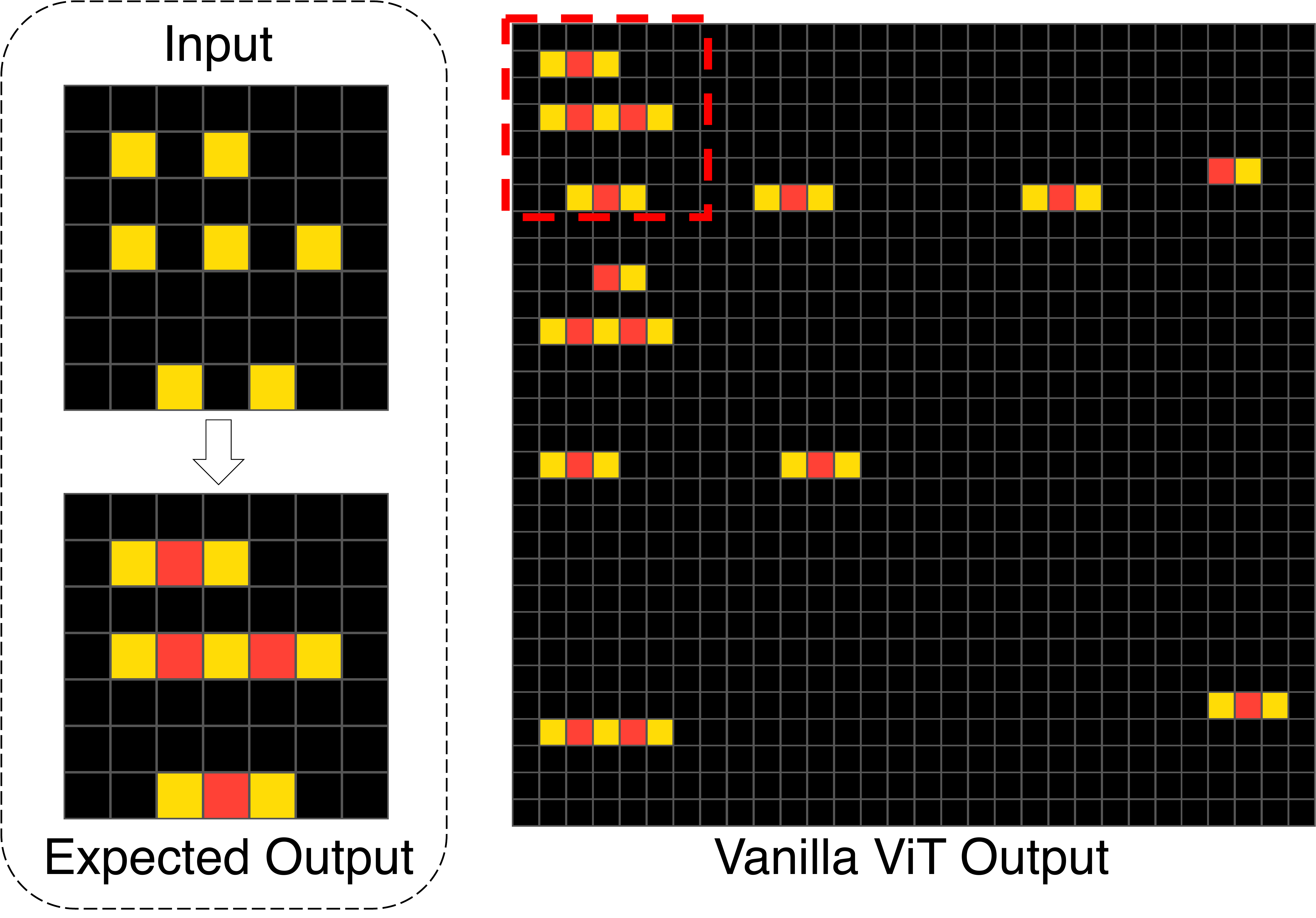}
\end{center}
\caption{\textbf{Failure case of ViT-Vanilla with NLP \texttt{<pad>} tokens.} ViT-Vanilla with 2D padding and NLP \texttt{<pad>} tokens fails to account for the actual inner grid size, filling the entire \( h_\text{max} \times w_\text{max} \) space. When the output is cropped to the true grid dimensions, the predictions within the valid region are correct, underscoring the importance of proper boundary handling.}

\label{fig:defaulViT_failure}
\end{figure}

\section{Training Details}
\label{appendix:training_detail}
This section provides a comprehensive overview of the training setup, including hyperparameters, hardware specifications, and other relevant details regarding the training process.

Our model consists of 3 layers with 8 attention heads and a hidden dimension of 128. The model was trained on various single-core GPU nodes, including P100, V100, and T4, with a batch size of 8 for 1 epoch. The typical training time per task ranges from 6 to 10 hours (wall clock).

The dataset was generated using Hodel's generators~\citep{hodel_rearc_2024}, producing 1 million samples, which were then split into training, validation, and test sets with 998,000, 1,000, and 1,000 instances, respectively. The generation time varies between 3 and 12 hours, depending on the task. A fixed random seed (1230) was used for both dataset generation and model training to ensure reproducibility.

Due to computational resource constraints, the ablation study was performed on a randomly sampled subset of 100 tasks from the total 400, also selected using seed 1230.

\subsection{Data-Regime Ablation}
\label{appendix:data_ablation}

To assess how our method scales with varying amounts of training data, we conducted an additional experiment on a random selection of 10 ARC tasks. We trained our final model (\method) from scratch using 1k, 10k, 50k, 100k, 500k, and 1M examples per task. Table~\ref{tab:data_ablation_results} summarizes the mean, standard deviation, and range of solve rates across these tasks.

\begin{table}[h]
\centering
\caption{Solved Test Instances (\%) on 10 ARC Tasks with Varying Training Set Sizes.}
\label{tab:data_ablation_results}
\begin{tabular}{r|rrrr}
\toprule
\textbf{Train Size} & \textbf{Mean} & \textbf{Std} & \textbf{Min} & \textbf{Max} \\
\midrule
1k    & 0.0000  & 0.0000   & 0.000 & 0.000 \\
10k   & 0.0000  & 0.0000   & 0.000 & 0.000 \\
50k   & 0.0001  & 0.0003   & 0.000 & 0.001 \\
100k  & 0.0010  & 0.0019   & 0.000 & 0.005 \\
500k  & 0.0046  & 0.0108   & 0.000 & 0.035 \\
1M    & 0.7314  & 0.4194   & 0.000 & 1.000 \\
\bottomrule
\end{tabular}
\end{table}

When the training set size falls below roughly 500k examples, performance on these 10 tasks remains near zero, regardless of task difficulty. Our model’s capacity (approximately 2.8M trainable parameters) appears to be the primary reason for needing around 1M training samples per task, as even 500k examples are insufficient to achieve non-trivial accuracy.




\section{PEmixer Strategies}
\label{appendix:pemixer_strategies}

In the main text (Section~\ref{sec:PositionalEnhancements}), we introduced the \emph{Positional Encoding Mixer (PEmixer)} as a straightforward yet effective method for balancing input embeddings and positional embeddings. Formally, we learn dimension-wise weights \(\boldsymbol{\alpha}\) and \(\boldsymbol{\beta}\) such that
\begin{equation}
\label{eq:pemixer_vector}
    h_{i}^{0}
    \;=\;
    \boldsymbol{\alpha} \,\odot\, \mathbf{E}_{p_{i}}
    \;+\;
    \boldsymbol{\beta} \,\odot\, \mathbf{E}_{\text{pos}_{i}},
\end{equation}
where \(\mathbf{E}_{p_{i}}\) and \(\mathbf{E}_{\text{pos}_{i}}\) denote the input (content) embedding and positional embedding for token \(i\), respectively, and \(\odot\) is element-wise multiplication. In this appendix, we provide a broader survey of mixing strategies, including normalization-based, attention-based, and scalar-based methods, and compare their performance on 10 randomly selected ARC tasks.

\subsection{Comparing Mixing Variants}

Below are the main variants we tested, all of which combine \(\mathbf{E}_{p_{i}}\) (input) and \(\mathbf{E}_{\text{pos}_{i}}\) (position) in different ways. Let \(\alpha\) and \(\beta\) be \emph{scalars} in some cases and \emph{vectors} in others, and let \(\text{Norm}(\cdot)\) represent \(L_2\)- or layer normalization.

\begin{enumerate}[leftmargin=*]
    \item \textbf{Default (no scalars).}  
    \begin{equation}
    \label{eq:pemix_default}
    h_{i}^{0} \;=\; \mathbf{E}_{p_{i}} \;+\; \mathbf{E}_{\text{pos}_{i}}.
    \end{equation}

    \item \textbf{Hardcoded Normalization.}  
    \begin{equation}
    \label{eq:pemix_hardcodednorm}
    h_{i}^{0}
    \;=\;
    \text{Norm}\bigl(\mathbf{E}_{p_{i}}\bigr)
    \;+\;
    \text{Norm}\bigl(\mathbf{E}_{\text{pos}_{i}}\bigr).
    \end{equation}

    \item \textbf{Learnable Scaling (Scalar or Vector).}  
    \begin{equation}
    \label{eq:pemix_learnablescaling}
    h_{i}^{0}
    \;=\;
    \mathbf{E}_{p_{i}}
    \;+\;
    \alpha \,\mathbf{E}_{\text{pos}_{i}},
    \end{equation}
    where \(\alpha\in\mathbb{R}\) or \(\alpha\in\mathbb{R}^{d}\). (If \(\alpha\) is a vector, it is applied element-wise.)

    \item \textbf{Weighted Sum (with or without normalization).}
    \begin{equation}
    \label{eq:pemix_weightedsum}
    h_{i}^{0}
    \;=\;
    \alpha_{\mathrm{in}}\; \text{Norm}\bigl(\mathbf{E}_{p_{i}}\bigr)
    \;+\;
    \beta_{\mathrm{pos}}\; \text{Norm}\bigl(\mathbf{E}_{\text{pos}_{i}}\bigr),
    \end{equation}
    or, in a version without normalization (``\_no\_norm''),
    \begin{equation}
    \label{eq:pemix_weightedsum_nonorm}
    h_{i}^{0}
    \;=\;
    \alpha_{\mathrm{in}}\;\mathbf{E}_{p_{i}}
    \;+\;
    \beta_{\mathrm{pos}}\;\mathbf{E}_{\text{pos}_{i}}.
    \end{equation}
    Here, \(\alpha_{\mathrm{in}}, \beta_{\mathrm{pos}}\in\mathbb{R}\) or \(\mathbb{R}^{d}\). (Again, if they are vectors, they apply element-wise.)

    \item \textbf{LayerNorm.}
    \begin{equation}
    \label{eq:pemix_layernorm}
    h_{i}^{0}
    \;=\;
    \text{LayerNorm}(\mathbf{E}_{p_{i}})
    \;+\;
    \text{LayerNorm}(\mathbf{E}_{\text{pos}_{i}}).
    \end{equation}

\end{enumerate}

We also include a \textbf{Learned APE (LAPE)} baseline that replaces the 2D sinusoidal embedding with a purely learnable positional embedding, holding all other components constant.

\subsection{Experimental Results}

In this subsection, we test a baseline configuration of ``ViTARC-VT + PEmixer'' (i.e., without OPE or 2DRPE), focusing specifically on different mixing strategies for combining content embeddings with positional embeddings. Table~\ref{tab:pemixer_variants} lists each strategy alongside its mean and median \emph{solved test-instance rate} over a sample of 10 ARC tasks. Our final choice, \texttt{weighted\_sum\_no\_norm\_vec}, corresponds to Equation~\eqref{eq:pemix_weightedsum_nonorm} where \(\alpha_{\mathrm{in}}, \beta_{\mathrm{pos}}\in\mathbb{R}^{d}\). It achieves the highest overall performance, slightly outperforming the \texttt{default (no scalars)} approach.

\begin{table}[h]
\centering
\caption{Performance of different PEmixer (or equivalent) strategies on 10 sampled ARC tasks.}
\label{tab:pemixer_variants}
\begin{tabular}{l|cc}
\toprule
\textbf{PEMix Variant} & \textbf{Mean} & \textbf{Median} \\
\midrule
\texttt{weighted\_sum\_no\_norm\_vec} & 0.7331 & 0.9780 \\
\texttt{default (no scalars)}         & 0.7179 & 0.9655 \\
\texttt{learnable\_scaling\_vec}      & 0.7088 & 0.8855 \\
\textbf{learnedAPE (LAPE)}            & 0.4006 & 0.2515 \\
\texttt{hardcoded\_normalization}     & 0.3128 & 0.1225 \\
\texttt{weighted\_sum\_vec}           & 0.1149 & 0.0000 \\
\texttt{learnable\_scaling}           & 0.0076 & 0.0010 \\
\texttt{layer\_norm}                  & 0.0044 & 0.0000 \\
\texttt{weighted\_sum\_no\_norm}      & 0.0000 & 0.0000 \\
\texttt{weighted\_sum}                & 0.0000 & 0.0000 \\
\bottomrule
\end{tabular}
\end{table}

\subsection{Task-Level Analysis and Learned PEmixer Weights}

Table~\ref{tab:per_task_comparison} compares \texttt{weighted\_sum\_no\_norm\_vec} to \texttt{default (no scalars)} on each of the 10 tasks, along with \(\textit{pos\_input\_ratio}\) = \(\beta_{\mathrm{pos}} / \alpha_{\mathrm{in}}\) averaged over embedding dimensions. When this ratio exceeds 1, the model places greater emphasis on positional information.

\begin{table}[h]
\centering
\caption{Per-task comparison of \texttt{default} vs. \texttt{weighted\_sum\_no\_norm\_vec}. 
“Diff” indicates absolute improvement, while \(\textit{pos\_input\_ratio}\) is the dimension-wise average ratio 
\(\beta_{\mathrm{pos}} / \alpha_{\mathrm{in}}\).}
\label{tab:per_task_comparison}
\begin{tabular}{l|cccc}
\toprule
\textbf{Task ID} & \textbf{Default} & \textbf{WS\_NoNorm\_Vec} & \textbf{Diff} & \(\textit{pos\_input\_ratio}\) \\
\midrule
b7249182 & 0.758 & 0.844 & 0.086 & 1.219 \\
5c2c9af4 & 0.495 & 0.525 & 0.030 & 1.249 \\
c444b776 & 0.965 & 0.980 & 0.015 & 1.261 \\
d89b689b & 0.966 & 0.976 & 0.010 & 1.145 \\
2bee17df & 0.994 & 0.999 & 0.005 & 1.398 \\
9af7a82c & 0.001 & 0.005 & 0.004 & 0.920 \\
c8cbb738 & 0.002 & 0.006 & 0.004 & 0.931 \\
ac0a08a4 & 0.999 & 1.000 & 0.001 & 1.132 \\
913fb3ed & 0.999 & 0.998 & -0.001 & 1.295 \\
d90796e8 & 1.000 & 0.998 & -0.002 & 1.345 \\
\bottomrule
\end{tabular}
\end{table}

\noindent
\textbf{Insights.} As shown in Table~\ref{tab:per_task_comparison}, most tasks see performance gains when 
\(\textit{pos\_input\_ratio}\) exceeds 1, indicating stronger emphasis on positional cues. In contrast, tasks with 
\(\textit{pos\_input\_ratio} < 1\) show only marginal gains (or minor drops). This corroborates our hypothesis 
that \emph{boosting positional information} often proves pivotal for abstract visual reasoning, particularly 
in tasks requiring complex spatial transformations.

\begin{figure}[h!]
    \centering
    \includegraphics[width=0.98\linewidth]{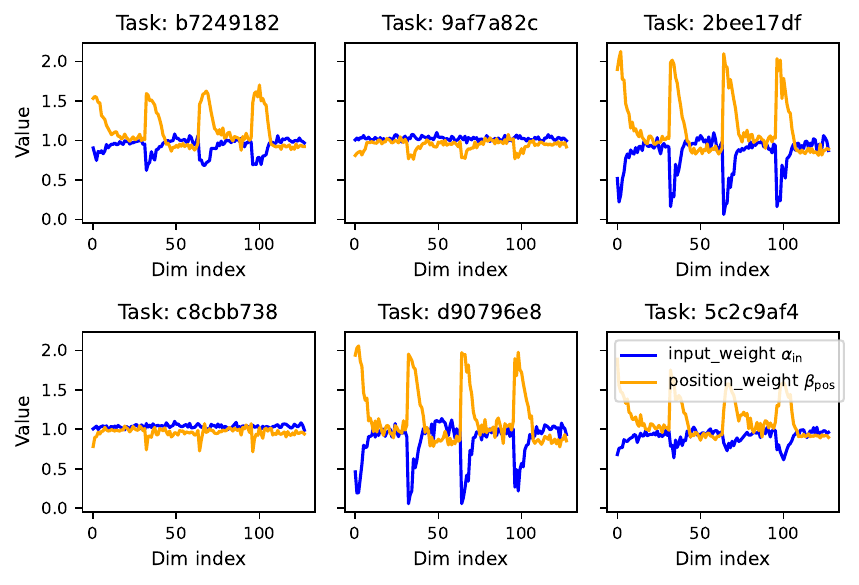}
    \caption{\textbf{Dimension-wise learned weights} for \texttt{weighted\_sum\_no\_norm\_vec} on 6 demo tasks. 
    Orange lines show the positional weights (\(\beta_{\mathrm{pos}}\)) and blue lines show the input weights (\(\alpha_{\mathrm{in}}\)). 
    Notice the periodic spikes aligned with sinusoidal cycles.}
    \label{fig:pemixer_scalars}
\end{figure}

Figure~\ref{fig:pemixer_scalars} visualizes how \(\beta_{\mathrm{pos}}\) (orange) and \(\alpha_{\mathrm{in}}\) (blue) vary across embedding dimensions. Several tasks exhibit a clear periodic pattern (around 32 channels), reflecting alignment with the sinusoidal encoding’s frequency structure. This indicates the network actively adjusts dimension-specific gains to emphasize critical spatial information.

\begin{figure}[h!]
    \centering
    \includegraphics[width=0.95\linewidth]{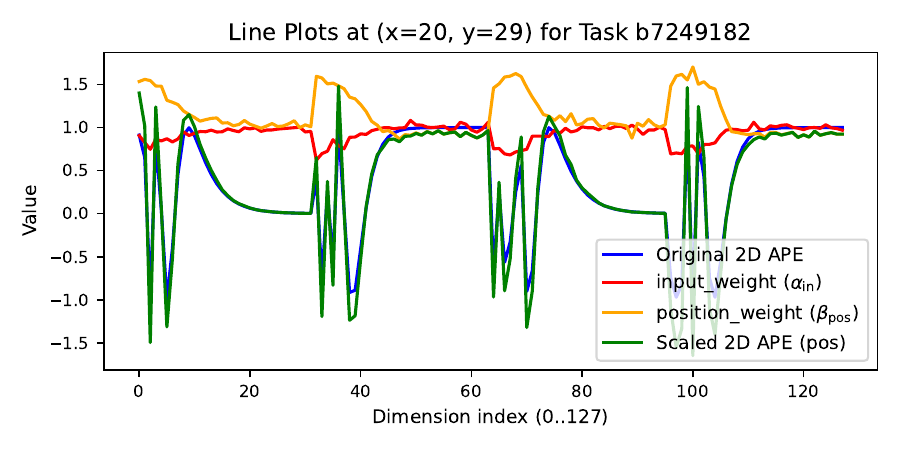}
    \caption{\textbf{Per-coordinate amplitudes for 2D APE, before and after applying the PEmixer scalars} 
    (\(\alpha_{\text{in}}\) for the input embedding and \(\beta_{\text{pos}}\) for the positional embedding). 
    Shown here is Task \texttt{b7249182}. We display the original sinusoidal embedding (top) for selected \((x,y)\) 
    coordinates, alongside the post-amplification embedding (bottom). Notice how lower-amplitude values 
    in the sine/cosine cycles receive a proportionally larger boost from \(\beta_{\text{pos}}\), while 
    \(\alpha_{\text{in}}\) simultaneously scales down the content embedding. This selective emphasis 
    makes \emph{positional} cues stand out more clearly, underscoring our hypothesis that \emph{boosting 
    positional information} is crucial for abstract visual reasoning.}
    \label{fig:b7249182_lineplot}
\end{figure}

\noindent
We then further illustrate how \texttt{weighted\_sum\_no\_norm\_vec} modifies raw 2D APE for specific pixel coordinates 
in Figure~\ref{fig:b7249182_lineplot} (Task \texttt{b7249182}). The top portion of the plot shows the unmodified 
sinusoidal wave, while the bottom portion shows the \emph{after-amplification} result, reflecting the 
element-wise multiplications by \(\alpha_{\text{in}}\) and \(\beta_{\text{pos}}\). We observe that the learned 
scalar tends to \emph{magnify} lower-scale parts of the sinusoidal wave, effectively ensuring certain spatial 
dimensions are more prominent. Meanwhile, the input embedding is suppressed, allowing positional structure 
to dominate where necessary. This behavior further validates our conclusion that dimension-wise weighting 
is vital for spatially-focused ARC tasks.

\subsection{Why Not Simply Use a Learned 2D APE (LAPE)?}

From Table~\ref{tab:pemixer_variants}, \textbf{learnedAPE} (LAPE) yields a mean solve rate of 0.4006—substantially lower than our best (0.7331). We hypothesize that a purely learned embedding lacks the strong inductive bias of sinusoidal geometry, which is particularly valuable in ARC tasks where pixel-level transformations are closely tied to grid coordinates. By combining the sinusoidal prior with dimension-wise learnable mixing, we preserve a stable spatial reference while allowing the model to amplify or downweight select dimensions as needed for each task.

\section{Additional 2D RPE Experiments}
\label{appendix:appendix_rpe}

In this section, we present further details on the 2D relative positional encoding (2D RPE) variants introduced in the main text. We evaluate them on a randomly selected subset of 50 tasks from the original 400-task ARC benchmark and summarize their performance in Table~\ref{tab:sample50_table}.

\subsection{Implementation Details}

\paragraph{(1) \textbf{ViTARC (2 directions)}} 
This baseline version extends the 1D ALiBi concept into 2D by grouping pixel positions ``above or to the left'' (assigned a slope \(r_{\text{before}}\)) versus ``below or to the right'' (assigned a slope \(r_{\text{after}}\)). 
Specifically, the 2D Manhattan distance 
\(d((x_i, y_i), (x_j, y_j))\)
between pixel coordinates \((x_i, y_i)\) and \((x_j, y_j)\) is multiplied by 
\(r_{\text{before}}\) if \(j \le i\), 
and by \(r_{\text{after}}\) if \(j > i\).
The slope values follow the geometric progression 
\(2^{-8/n}\) for \(n\) heads, 
with \(r_{\text{before}}\) initialized at \(1/2^1\) and \(r_{\text{after}}\) at \(1/2^{0.5}\).

\paragraph{(2) \textbf{2D RPE -- Four Diagonal Directions}}
This version splits the encoding into four diagonal directions, assigning distinct slopes 
\(\{r_{\text{tl}}, r_{\text{tr}}, r_{\text{dl}}, r_{\text{dr}}\}\)
to top-left, top-right, down-left, and down-right, respectively. 
As in the two-direction version, we multiply the 2D Manhattan distance by the appropriate slope based on whether \(\mathbf{P}_j\) lies diagonally above-left, above-right, below-left, or below-right of \(\mathbf{P}_i\). 
Example initializations include
\[
  r_{\text{tl}} = \tfrac{1}{2^{0.25}},\quad
  r_{\text{tr}} = \tfrac{1}{2},\quad
  r_{\text{dl}} = \tfrac{1}{2^{0.75}},\quad
  r_{\text{dr}} = \tfrac{1}{2^{0.5}}.
\]

\paragraph{(3) \textbf{2D RPE -- Four Directions (Cardinal)}}
In this variant, we use four distinct slopes for pixels above, below, left, or right, ignoring diagonal positioning. 
Conceptually, each direction (up, down, left, right) has its own slope, and the total bias is scaled accordingly for pixel pairs that differ primarily along one axis. 
For example, positions that are directly above get multiplied by the \emph{up} slope, while positions directly to the left are scaled by the \emph{left} slope, etc.

\subsection{Performance on 50 ARC Tasks}

Table~\ref{tab:sample50_table} summarizes the mean, median, and percentile performance across the three 2D RPE configurations, evaluated on a 50-task subset:

\begin{table}[htbp]
\centering
\resizebox{\textwidth}{!}{
\begin{tabular}{l|lllll}
\toprule
\multirow{2}{*}{Model (on 50 subset)} & \multicolumn{4}{c}{Solved Test Instances (\%)} & \multirow{2}{*}{\textbf{Delta (Mean)}} \\ 
 & \multicolumn{1}{l}{Mean} & \multicolumn{1}{l}{Median} & \multicolumn{1}{l}{25th Pctl.} & \multicolumn{1}{l}{75th Pctl.} & \\ \midrule
ViTARC (2 directions)        & \multicolumn{1}{l}{79.51} & \multicolumn{1}{l}{\textbf{96.90}} & \multicolumn{1}{l}{\textbf{83.12}} & \textbf{99.88} & \multicolumn{1}{l}{base} \\ 
- 2D RPE - 4 diag directions & \multicolumn{1}{l}{\textbf{79.74}} & \multicolumn{1}{l}{94.20} & \multicolumn{1}{l}{77.72} & 99.80 & \multicolumn{1}{l}{+0.23} \\
- 2D RPE - 4 directions (cardinal) & \multicolumn{1}{l}{78.10} & \multicolumn{1}{l}{96.50} & \multicolumn{1}{l}{71.70} & 99.83 & \multicolumn{1}{l}{-1.41} \\
\bottomrule
\end{tabular}
}
\caption{Performance comparison of different 2D RPE variants on a 50-task ARC subset.}
\label{tab:sample50_table}
\end{table}

\subsection{Insights}
The results indicate that the four-diagonal-direction RPE achieves a slight improvement in mean performance (+0.23) over the two-direction baseline, suggesting that explicitly modeling diagonal relationships enhances spatial encoding for certain tasks. However, while the four-slope approach produces a higher overall mean, the two-slope version shows notable advantages in median, 25th percentile, and 75th percentile performance. This suggests that incorporating diagonal relationships is beneficial, yet the finer granularity of four distinct diagonal slopes may not always serve as a robust inductive bias across diverse tasks; it can substantially boost performance on some problems, but does not necessarily generalize as effectively as the simpler two-slope design. Meanwhile, both Manhattan-based configurations generally outperform the strictly cardinal-based approach in mean performance, reinforcing the notion that diagonals capture valuable spatial cues critical for tasks involving complex pattern recognition.


\section{Full Results for Task-specific Accuracies}

\subsection{Main models on full 400 tasks}
\label{appendix:full_results}

\begin{table}[!htbp]
\centering
\label{tab:full400_table}

\caption{Solved Test Instances (\%) Across Models on all 400 tasks.}
\begin{tabular}{l|llll}
\toprule
\multirow{2}{*}{Model} & \multicolumn{4}{c}{Solved Test Instances (\%)} \\ 
                 & \multicolumn{1}{l|}{Mean} & \multicolumn{1}{l|}{Med.} & \multicolumn{1}{l|}{25th Pctl.} & 75th Pctl. \\ \midrule
Baseline (ViT-Vanilla)     & \multicolumn{1}{l|}{17.68} & \multicolumn{1}{l|}{3.20} & \multicolumn{1}{l|}{0.10} & 22.85 \\
ViTARC-VT                 & \multicolumn{1}{l|}{66.03} & \multicolumn{1}{l|}{87.85} & \multicolumn{1}{l|}{27.55} & 99.30 \\
ViTARC (Full Model)        & \multicolumn{1}{l|}{75.04} & \multicolumn{1}{l|}{95.10} & \multicolumn{1}{l|}{58.07} & 99.80 \\ 
\bottomrule
\end{tabular}
\end{table}

\begin{table}[!htbp]
\centering
\caption{Model accuracies across tasks (100/400)}
\begin{tabular}{c|ccc!{\vrule width 1pt}c|ccc}
\toprule
Task & ViT & ViTARC & ViTARC & Task & ViT & ViTARC & ViTARC \\
 & -Vanilla & -VT &  &  & -Vanilla & -VT &  \\
\midrule
ce22a75a & 0.00 & 0.94 & 1.00 & 444801d8 & 0.00 & 0.98 & 1.00 \\
1f876c06 & 0.00 & 0.99 & 1.00 & b27ca6d3 & 0.00 & 0.99 & 1.00 \\
68b16354 & 0.00 & 0.99 & 1.00 & 2c608aff & 0.00 & 1.00 & 1.00 \\
d037b0a7 & 0.00 & 1.00 & 1.00 & 0ca9ddb6 & 0.00 & 1.00 & 1.00 \\
543a7ed5 & 0.00 & 1.00 & 1.00 & 952a094c & 0.00 & 1.00 & 1.00 \\
af902bf9 & 0.00 & 1.00 & 1.00 & 49d1d64f & 0.00 & 1.00 & 1.00 \\
0962bcdd & 0.00 & 1.00 & 1.00 & d364b489 & 0.00 & 1.00 & 1.00 \\
b60334d2 & 0.00 & 1.00 & 1.00 & a9f96cdd & 0.00 & 1.00 & 1.00 \\
95990924 & 0.00 & 1.00 & 1.00 & 54d82841 & 0.00 & 0.80 & 0.99 \\
25d487eb & 0.00 & 0.95 & 0.99 & 5c0a986e & 0.00 & 0.96 & 0.99 \\
d687bc17 & 0.00 & 0.97 & 0.99 & 363442ee & 0.00 & 0.98 & 0.99 \\
6cdd2623 & 0.00 & 0.98 & 0.99 & db93a21d & 0.00 & 0.93 & 0.97 \\
5168d44c & 0.00 & 0.94 & 0.97 & 3befdf3e & 0.00 & 0.97 & 0.97 \\
22233c11 & 0.00 & 0.97 & 0.97 & 67a3c6ac & 0.00 & 1.00 & 0.97 \\
ae3edfdc & 0.00 & 0.72 & 0.96 & ded97339 & 0.00 & 0.92 & 0.96 \\
a2fd1cf0 & 0.00 & 0.95 & 0.96 & d4a91cb9 & 0.00 & 0.98 & 0.96 \\
d4f3cd78 & 0.00 & 0.99 & 0.96 & 6cf79266 & 0.00 & 0.96 & 0.95 \\
e98196ab & 0.00 & 0.99 & 0.95 & 56ff96f3 & 0.00 & 0.90 & 0.94 \\
694f12f3 & 0.00 & 0.91 & 0.94 & 93b581b8 & 0.00 & 0.99 & 0.94 \\
39e1d7f9 & 0.00 & 0.42 & 0.93 & 8403a5d5 & 0.00 & 1.00 & 0.93 \\
ecdecbb3 & 0.00 & 0.76 & 0.92 & 31aa019c & 0.00 & 0.82 & 0.90 \\
ec883f72 & 0.00 & 0.87 & 0.90 & 36fdfd69 & 0.00 & 0.75 & 0.89 \\
b7249182 & 0.00 & 0.74 & 0.88 & e9614598 & 0.00 & 0.86 & 0.88 \\
e76a88a6 & 0.00 & 0.00 & 0.87 & 3ac3eb23 & 0.00 & 0.71 & 0.87 \\
a64e4611 & 0.00 & 0.98 & 0.87 & 50846271 & 0.00 & 0.84 & 0.86 \\
928ad970 & 0.00 & 0.97 & 0.86 & 40853293 & 0.00 & 0.99 & 0.86 \\
6ecd11f4 & 0.00 & 0.00 & 0.84 & b527c5c6 & 0.00 & 0.66 & 0.84 \\
1e0a9b12 & 0.00 & 0.69 & 0.84 & 7ddcd7ec & 0.00 & 0.75 & 0.84 \\
2013d3e2 & 0.00 & 0.95 & 0.84 & e50d258f & 0.00 & 0.70 & 0.83 \\
1caeab9d & 0.00 & 0.42 & 0.82 & 5ad4f10b & 0.00 & 0.62 & 0.82 \\
98cf29f8 & 0.00 & 0.66 & 0.82 & 264363fd & 0.00 & 0.79 & 0.82 \\
5521c0d9 & 0.00 & 0.75 & 0.79 & 0a938d79 & 0.00 & 0.86 & 0.78 \\
f8a8fe49 & 0.00 & 0.68 & 0.74 & a48eeaf7 & 0.00 & 0.76 & 0.73 \\
aba27056 & 0.00 & 0.59 & 0.70 & 2bcee788 & 0.00 & 0.64 & 0.70 \\
47c1f68c & 0.00 & 0.45 & 0.68 & b548a754 & 0.00 & 0.95 & 0.68 \\
890034e9 & 0.00 & 0.59 & 0.67 & 508bd3b6 & 0.00 & 0.66 & 0.64 \\
6aa20dc0 & 0.00 & 0.33 & 0.63 & 2dd70a9a & 0.00 & 0.33 & 0.59 \\
7c008303 & 0.00 & 0.48 & 0.58 & 6d58a25d & 0.00 & 0.33 & 0.56 \\
f8c80d96 & 0.00 & 0.13 & 0.55 & 6855a6e4 & 0.00 & 0.44 & 0.51 \\
4093f84a & 0.00 & 0.31 & 0.49 & 90c28cc7 & 0.00 & 0.42 & 0.48 \\
db3e9e38 & 0.00 & 0.34 & 0.47 & 05f2a901 & 0.00 & 0.04 & 0.46 \\
5c2c9af4 & 0.00 & 0.51 & 0.46 & d06dbe63 & 0.00 & 0.57 & 0.46 \\
5daaa586 & 0.00 & 0.17 & 0.43 & f1cefba8 & 0.00 & 0.19 & 0.43 \\
3906de3d & 0.00 & 0.28 & 0.42 & caa06a1f & 0.00 & 0.19 & 0.41 \\
75b8110e & 0.00 & 0.62 & 0.40 & e8dc4411 & 0.00 & 0.28 & 0.39 \\
8731374e & 0.00 & 0.22 & 0.38 & e48d4e1a & 0.00 & 0.30 & 0.38 \\
f35d900a & 0.00 & 0.65 & 0.38 & f15e1fac & 0.00 & 0.10 & 0.37 \\
6e19193c & 0.00 & 0.12 & 0.37 & 3de23699 & 0.00 & 0.00 & 0.35 \\
6b9890af & 0.00 & 0.00 & 0.35 & a78176bb & 0.00 & 0.26 & 0.32 \\
1b60fb0c & 0.00 & 0.14 & 0.28 & e509e548 & 0.00 & 0.02 & 0.27 \\
\bottomrule\end{tabular}
\end{table}

\begin{table}[htbp]
\centering
\caption{Model accuracies across tasks (200/400)}
\begin{tabular}{c|ccc!{\vrule width 1pt}c|ccc}
\toprule
Task & ViT & ViTARC & ViTARC & Task & ViT & ViTARC & ViTARC \\
 & -Vanilla & -VT &  &  & -Vanilla & -VT &  \\
\midrule
a1570a43 & 0.00 & 0.54 & 0.25 & 3e980e27 & 0.00 & 0.02 & 0.22 \\
88a10436 & 0.00 & 0.00 & 0.20 & 9aec4887 & 0.00 & 0.02 & 0.19 \\
7df24a62 & 0.00 & 0.10 & 0.19 & e21d9049 & 0.00 & 0.10 & 0.19 \\
8a004b2b & 0.00 & 0.02 & 0.18 & 1f0c79e5 & 0.00 & 0.14 & 0.16 \\
045e512c & 0.00 & 0.06 & 0.14 & ce602527 & 0.00 & 0.00 & 0.12 \\
b775ac94 & 0.00 & 0.03 & 0.12 & 8eb1be9a & 0.00 & 0.03 & 0.07 \\
fcb5c309 & 0.00 & 0.00 & 0.06 & a61ba2ce & 0.00 & 0.00 & 0.06 \\
36d67576 & 0.00 & 0.04 & 0.06 & 846bdb03 & 0.00 & 0.00 & 0.05 \\
234bbc79 & 0.00 & 0.00 & 0.05 & e40b9e2f & 0.00 & 0.02 & 0.05 \\
57aa92db & 0.00 & 0.03 & 0.05 & 5117e062 & 0.00 & 0.00 & 0.04 \\
8efcae92 & 0.00 & 0.00 & 0.04 & 72322fa7 & 0.00 & 0.02 & 0.04 \\
623ea044 & 0.00 & 0.02 & 0.04 & 4938f0c2 & 0.00 & 0.07 & 0.04 \\
3bd67248 & 0.00 & 0.08 & 0.04 & 48d8fb45 & 0.00 & 0.00 & 0.03 \\
a87f7484 & 0.00 & 0.00 & 0.03 & 447fd412 & 0.00 & 0.01 & 0.03 \\
e6721834 & 0.00 & 0.01 & 0.03 & 4c5c2cf0 & 0.00 & 0.08 & 0.03 \\
be94b721 & 0.00 & 0.00 & 0.02 & a8c38be5 & 0.00 & 0.00 & 0.02 \\
d07ae81c & 0.00 & 0.00 & 0.01 & 97a05b5b & 0.00 & 0.01 & 0.01 \\
99b1bc43 & 0.00 & 0.00 & 0.00 & 137eaa0f & 0.00 & 0.00 & 0.00 \\
c8cbb738 & 0.00 & 0.00 & 0.00 & e5062a87 & 0.00 & 0.00 & 0.00 \\
60b61512 & 0.01 & 0.83 & 1.00 & e8593010 & 0.01 & 0.83 & 1.00 \\
a79310a0 & 0.01 & 0.98 & 1.00 & d43fd935 & 0.01 & 0.98 & 1.00 \\
253bf280 & 0.01 & 0.99 & 1.00 & dbc1a6ce & 0.01 & 1.00 & 1.00 \\
4c4377d9 & 0.01 & 1.00 & 1.00 & 8be77c9e & 0.01 & 1.00 & 1.00 \\
77fdfe62 & 0.01 & 1.00 & 1.00 & ed36ccf7 & 0.01 & 1.00 & 1.00 \\
25ff71a9 & 0.01 & 1.00 & 1.00 & f5b8619d & 0.01 & 1.00 & 1.00 \\
dc1df850 & 0.01 & 1.00 & 1.00 & 10fcaaa3 & 0.01 & 0.99 & 0.99 \\
178fcbfb & 0.01 & 1.00 & 0.99 & 3428a4f5 & 0.01 & 0.79 & 0.98 \\
11852cab & 0.01 & 0.92 & 0.98 & 4612dd53 & 0.01 & 0.96 & 0.98 \\
fcc82909 & 0.01 & 0.96 & 0.97 & dc433765 & 0.01 & 0.91 & 0.96 \\
39a8645d & 0.01 & 0.01 & 0.94 & 6fa7a44f & 0.01 & 1.00 & 0.94 \\
834ec97d & 0.01 & 0.94 & 0.93 & 321b1fc6 & 0.01 & 0.55 & 0.92 \\
4522001f & 0.01 & 0.22 & 0.88 & 88a62173 & 0.01 & 0.97 & 0.85 \\
d9f24cd1 & 0.01 & 0.67 & 0.74 & a65b410d & 0.01 & 0.69 & 0.74 \\
9edfc990 & 0.01 & 0.33 & 0.48 & 6455b5f5 & 0.01 & 0.22 & 0.27 \\
72ca375d & 0.01 & 0.01 & 0.14 & 3f7978a0 & 0.01 & 0.04 & 0.14 \\
f9012d9b & 0.01 & 0.02 & 0.02 & 0e206a2e & 0.01 & 0.02 & 0.02 \\
a8d7556c & 0.02 & 0.93 & 1.00 & 74dd1130 & 0.02 & 1.00 & 1.00 \\
d13f3404 & 0.02 & 1.00 & 1.00 & 6d0aefbc & 0.02 & 1.00 & 1.00 \\
c9e6f938 & 0.02 & 1.00 & 1.00 & 913fb3ed & 0.02 & 1.00 & 1.00 \\
41e4d17e & 0.02 & 0.83 & 0.99 & 94f9d214 & 0.02 & 0.74 & 0.96 \\
83302e8f & 0.02 & 0.75 & 0.94 & b94a9452 & 0.02 & 0.45 & 0.85 \\
1f85a75f & 0.02 & 0.03 & 0.81 & b6afb2da & 0.02 & 1.00 & 0.77 \\
6e82a1ae & 0.02 & 0.24 & 0.63 & 00d62c1b & 0.02 & 0.46 & 0.63 \\
82819916 & 0.02 & 0.20 & 0.60 & 63613498 & 0.02 & 0.02 & 0.16 \\
228f6490 & 0.02 & 0.03 & 0.06 & 09629e4f & 0.02 & 0.02 & 0.03 \\
6d75e8bb & 0.03 & 0.99 & 1.00 & bc1d5164 & 0.03 & 1.00 & 1.00 \\
bdad9b1f & 0.03 & 1.00 & 1.00 & eb281b96 & 0.03 & 1.00 & 1.00 \\
e26a3af2 & 0.03 & 0.92 & 0.99 & 8d510a79 & 0.03 & 0.99 & 0.99 \\
f2829549 & 0.03 & 0.89 & 0.98 & 6430c8c4 & 0.03 & 0.89 & 0.98 \\
f25fbde4 & 0.03 & 0.02 & 0.96 & fafffa47 & 0.03 & 0.92 & 0.94 \\
\bottomrule\end{tabular}
\end{table}

\begin{table}[htbp]
\centering
\caption{Model accuracies across tasks (300/400)}
\begin{tabular}{c|ccc!{\vrule width 1pt}c|ccc}
\toprule
Task & ViT & ViTARC & ViTARC & Task & ViT & ViTARC & ViTARC \\
 & -Vanilla & -VT &  &  & -Vanilla & -VT &  \\
\midrule
6773b310 & 0.03 & 0.78 & 0.91 & a740d043 & 0.03 & 0.84 & 0.84 \\
56dc2b01 & 0.03 & 0.43 & 0.58 & d2abd087 & 0.03 & 0.09 & 0.15 \\
681b3aeb & 0.03 & 0.04 & 0.05 & 5bd6f4ac & 0.04 & 1.00 & 1.00 \\
8d5021e8 & 0.04 & 1.00 & 1.00 & 3c9b0459 & 0.04 & 1.00 & 1.00 \\
6150a2bd & 0.04 & 1.00 & 1.00 & 62c24649 & 0.04 & 1.00 & 0.99 \\
3af2c5a8 & 0.04 & 1.00 & 0.99 & 1a07d186 & 0.04 & 0.84 & 0.98 \\
855e0971 & 0.04 & 0.96 & 0.98 & 4258a5f9 & 0.04 & 0.97 & 0.98 \\
3aa6fb7a & 0.04 & 1.00 & 0.98 & 6d0160f0 & 0.04 & 0.03 & 0.97 \\
29ec7d0e & 0.04 & 0.62 & 0.83 & ae4f1146 & 0.04 & 0.14 & 0.67 \\
760b3cac & 0.04 & 0.66 & 0.64 & 29623171 & 0.04 & 0.37 & 0.44 \\
673ef223 & 0.04 & 0.30 & 0.26 & 2281f1f4 & 0.05 & 1.00 & 1.00 \\
cf98881b & 0.05 & 1.00 & 1.00 & ce4f8723 & 0.05 & 0.97 & 0.99 \\
6c434453 & 0.05 & 0.93 & 0.96 & c1d99e64 & 0.05 & 0.99 & 0.95 \\
2dc579da & 0.05 & 0.38 & 0.69 & c909285e & 0.05 & 0.20 & 0.58 \\
73251a56 & 0.05 & 0.66 & 0.39 & 776ffc46 & 0.05 & 0.03 & 0.16 \\
3345333e & 0.05 & 0.08 & 0.14 & beb8660c & 0.05 & 0.09 & 0.09 \\
80af3007 & 0.06 & 0.98 & 1.00 & 7f4411dc & 0.06 & 0.95 & 0.99 \\
32597951 & 0.06 & 0.98 & 0.99 & 7468f01a & 0.06 & 0.42 & 0.84 \\
810b9b61 & 0.06 & 0.70 & 0.82 & a5313dff & 0.06 & 0.61 & 0.76 \\
ef135b50 & 0.07 & 0.99 & 1.00 & dae9d2b5 & 0.07 & 0.95 & 0.97 \\
1c786137 & 0.07 & 0.05 & 0.75 & d8c310e9 & 0.07 & 0.72 & 0.74 \\
d22278a0 & 0.07 & 0.70 & 0.66 & d0f5fe59 & 0.08 & 0.09 & 1.00 \\
d5d6de2d & 0.08 & 0.98 & 1.00 & a416b8f3 & 0.08 & 1.00 & 1.00 \\
1f642eb9 & 0.08 & 1.00 & 1.00 & c444b776 & 0.08 & 0.96 & 0.99 \\
cbded52d & 0.08 & 0.97 & 0.97 & 780d0b14 & 0.08 & 0.97 & 0.96 \\
0b148d64 & 0.08 & 0.26 & 0.62 & b782dc8a & 0.08 & 0.30 & 0.28 \\
9f236235 & 0.09 & 0.98 & 0.88 & 0dfd9992 & 0.09 & 0.67 & 0.84 \\
7837ac64 & 0.09 & 0.82 & 0.82 & aabf363d & 0.09 & 0.12 & 0.73 \\
b8cdaf2b & 0.09 & 0.64 & 0.61 & a61f2674 & 0.10 & 0.75 & 0.84 \\
ce9e57f2 & 0.10 & 0.75 & 0.83 & 7b6016b9 & 0.10 & 0.65 & 0.80 \\
0520fde7 & 0.11 & 1.00 & 1.00 & 496994bd & 0.11 & 1.00 & 0.97 \\
150deff5 & 0.11 & 0.91 & 0.95 & 25d8a9c8 & 0.12 & 0.46 & 1.00 \\
1b2d62fb & 0.12 & 0.99 & 1.00 & 1bfc4729 & 0.12 & 1.00 & 1.00 \\
3618c87e & 0.12 & 0.98 & 0.99 & 90f3ed37 & 0.12 & 0.83 & 0.84 \\
484b58aa & 0.12 & 0.54 & 0.66 & 662c240a & 0.12 & 0.77 & 0.42 \\
b2862040 & 0.12 & 0.30 & 0.39 & d90796e8 & 0.13 & 1.00 & 1.00 \\
6a1e5592 & 0.13 & 0.18 & 0.22 & 42a50994 & 0.14 & 0.98 & 1.00 \\
2bee17df & 0.14 & 0.99 & 1.00 & 67e8384a & 0.14 & 1.00 & 1.00 \\
017c7c7b & 0.14 & 0.95 & 0.99 & a3325580 & 0.14 & 0.01 & 0.00 \\
ddf7fa4f & 0.15 & 0.78 & 0.95 & 23b5c85d & 0.16 & 0.03 & 0.24 \\
05269061 & 0.16 & 0.12 & 0.22 & 22168020 & 0.17 & 1.00 & 1.00 \\
23581191 & 0.17 & 0.92 & 0.96 & 53b68214 & 0.17 & 0.94 & 0.96 \\
7e0986d6 & 0.18 & 0.97 & 1.00 & b190f7f5 & 0.18 & 0.97 & 0.98 \\
a3df8b1e & 0.18 & 0.14 & 0.22 & ea786f4a & 0.19 & 0.98 & 0.98 \\
28bf18c6 & 0.19 & 0.07 & 0.81 & 3eda0437 & 0.19 & 0.68 & 0.69 \\
22eb0ac0 & 0.20 & 0.96 & 1.00 & 3631a71a & 0.20 & 0.99 & 1.00 \\
aedd82e4 & 0.20 & 1.00 & 1.00 & 025d127b & 0.20 & 1.00 & 1.00 \\
08ed6ac7 & 0.20 & 0.99 & 0.95 & 44d8ac46 & 0.20 & 0.59 & 0.86 \\
ff805c23 & 0.21 & 0.10 & 0.28 & e179c5f4 & 0.22 & 0.01 & 0.01 \\
1cf80156 & 0.23 & 0.12 & 0.83 & f8ff0b80 & 0.23 & 0.33 & 0.65 \\
\bottomrule\end{tabular}
\end{table}

\begin{table}[htbp]
\centering
\caption{Model accuracies across tasks (400/400)}
\begin{tabular}{c|ccc!{\vrule width 1pt}c|ccc}
\toprule
Task & ViT & ViTARC & ViTARC & Task & ViT & ViTARC & ViTARC \\
 & -Vanilla & -VT &  &  & -Vanilla & -VT &  \\
\midrule
1fad071e & 0.23 & 0.24 & 0.59 & 9ecd008a & 0.23 & 0.16 & 0.24 \\
67385a82 & 0.24 & 1.00 & 1.00 & 868de0fa & 0.24 & 1.00 & 1.00 \\
c9f8e694 & 0.24 & 1.00 & 1.00 & d6ad076f & 0.24 & 0.98 & 0.99 \\
dc0a314f & 0.24 & 0.14 & 0.24 & 27a28665 & 0.26 & 0.24 & 0.94 \\
9af7a82c & 0.26 & 0.00 & 0.00 & 4290ef0e & 0.27 & 0.24 & 0.80 \\
539a4f51 & 0.28 & 0.72 & 0.76 & cdecee7f & 0.28 & 0.04 & 0.11 \\
99fa7670 & 0.29 & 1.00 & 1.00 & e73095fd & 0.29 & 0.98 & 0.99 \\
9dfd6313 & 0.29 & 0.99 & 0.99 & b0c4d837 & 0.29 & 0.21 & 0.97 \\
963e52fc & 0.30 & 1.00 & 1.00 & 941d9a10 & 0.30 & 0.98 & 0.99 \\
b230c067 & 0.30 & 0.44 & 0.46 & b9b7f026 & 0.31 & 0.37 & 1.00 \\
06df4c85 & 0.31 & 1.00 & 1.00 & 67a423a3 & 0.32 & 1.00 & 0.99 \\
54d9e175 & 0.33 & 1.00 & 1.00 & 28e73c20 & 0.33 & 1.00 & 0.98 \\
6f8cd79b & 0.33 & 1.00 & 0.98 & ea32f347 & 0.34 & 0.65 & 0.71 \\
97999447 & 0.35 & 1.00 & 1.00 & a85d4709 & 0.35 & 0.00 & 0.83 \\
a5f85a15 & 0.36 & 0.99 & 1.00 & c59eb873 & 0.36 & 1.00 & 1.00 \\
7b7f7511 & 0.36 & 0.89 & 0.95 & d10ecb37 & 0.39 & 1.00 & 1.00 \\
d89b689b & 0.41 & 0.96 & 0.98 & de1cd16c & 0.41 & 0.37 & 0.97 \\
29c11459 & 0.43 & 1.00 & 1.00 & 9172f3a0 & 0.43 & 1.00 & 1.00 \\
a68b268e & 0.44 & 1.00 & 1.00 & ba97ae07 & 0.44 & 1.00 & 1.00 \\
ff28f65a & 0.44 & 0.70 & 0.96 & 1190e5a7 & 0.44 & 0.81 & 0.91 \\
d406998b & 0.46 & 0.98 & 1.00 & ba26e723 & 0.47 & 1.00 & 1.00 \\
f25ffba3 & 0.50 & 0.99 & 1.00 & c3f564a4 & 0.52 & 0.94 & 1.00 \\
2204b7a8 & 0.52 & 0.96 & 0.98 & 272f95fa & 0.54 & 1.00 & 1.00 \\
91714a58 & 0.54 & 0.94 & 0.98 & 1e32b0e9 & 0.56 & 0.99 & 1.00 \\
d9fac9be & 0.57 & 0.68 & 0.97 & 44f52bb0 & 0.57 & 0.55 & 0.84 \\
d23f8c26 & 0.59 & 1.00 & 1.00 & b8825c91 & 0.60 & 0.99 & 0.99 \\
ac0a08a4 & 0.61 & 0.99 & 1.00 & bb43febb & 0.61 & 1.00 & 1.00 \\
c0f76784 & 0.61 & 1.00 & 1.00 & e9afcf9a & 0.62 & 1.00 & 0.98 \\
b91ae062 & 0.64 & 1.00 & 1.00 & cce03e0d & 0.64 & 1.00 & 1.00 \\
007bbfb7 & 0.65 & 0.99 & 1.00 & 91413438 & 0.65 & 0.38 & 0.32 \\
c3e719e8 & 0.66 & 0.99 & 1.00 & e3497940 & 0.66 & 1.00 & 1.00 \\
d631b094 & 0.66 & 0.41 & 0.64 & 50cb2852 & 0.68 & 1.00 & 1.00 \\
8e1813be & 0.70 & 0.99 & 1.00 & 9565186b & 0.74 & 0.96 & 1.00 \\
a699fb00 & 0.74 & 1.00 & 1.00 & 4347f46a & 0.76 & 1.00 & 0.99 \\
469497ad & 0.76 & 0.92 & 0.95 & 239be575 & 0.76 & 0.74 & 0.82 \\
8f2ea7aa & 0.81 & 0.23 & 0.98 & 5614dbcf & 0.82 & 1.00 & 1.00 \\
9d9215db & 0.83 & 0.96 & 0.97 & 85c4e7cd & 0.84 & 0.99 & 0.90 \\
8e5a5113 & 0.85 & 0.98 & 0.99 & 46442a0e & 0.86 & 1.00 & 1.00 \\
7fe24cdd & 0.86 & 1.00 & 1.00 & 445eab21 & 0.86 & 0.92 & 0.96 \\
bd4472b8 & 0.89 & 0.49 & 0.58 & 3bdb4ada & 0.92 & 1.00 & 1.00 \\
bda2d7a6 & 0.94 & 0.98 & 1.00 & f76d97a5 & 0.94 & 1.00 & 1.00 \\
2dee498d & 0.95 & 1.00 & 1.00 & 46f33fce & 0.96 & 1.00 & 1.00 \\
746b3537 & 0.96 & 0.99 & 0.99 & eb5a1d5d & 0.97 & 1.00 & 1.00 \\
0d3d703e & 0.98 & 1.00 & 1.00 & 5582e5ca & 0.98 & 0.95 & 0.99 \\
f8b3ba0a & 0.98 & 0.99 & 0.97 & feca6190 & 0.98 & 0.11 & 0.79 \\
794b24be & 0.98 & 0.24 & 0.23 & d511f180 & 0.99 & 1.00 & 1.00 \\
b1948b0a & 0.99 & 1.00 & 1.00 & c8f0f002 & 0.99 & 1.00 & 1.00 \\
995c5fa3 & 1.00 & 0.00 & 1.00 & 6e02f1e3 & 1.00 & 1.00 & 1.00 \\
bbc9ae5d & 1.00 & 1.00 & 1.00 & d4469b4b & 1.00 & 1.00 & 1.00 \\
7447852a & 1.00 & 1.00 & 1.00 & 4be741c5 & 1.00 & 1.00 & 1.00 \\
\bottomrule\end{tabular}
\end{table}

\clearpage

\begin{figure}
    \centering
    \includegraphics[width=0.8\linewidth]{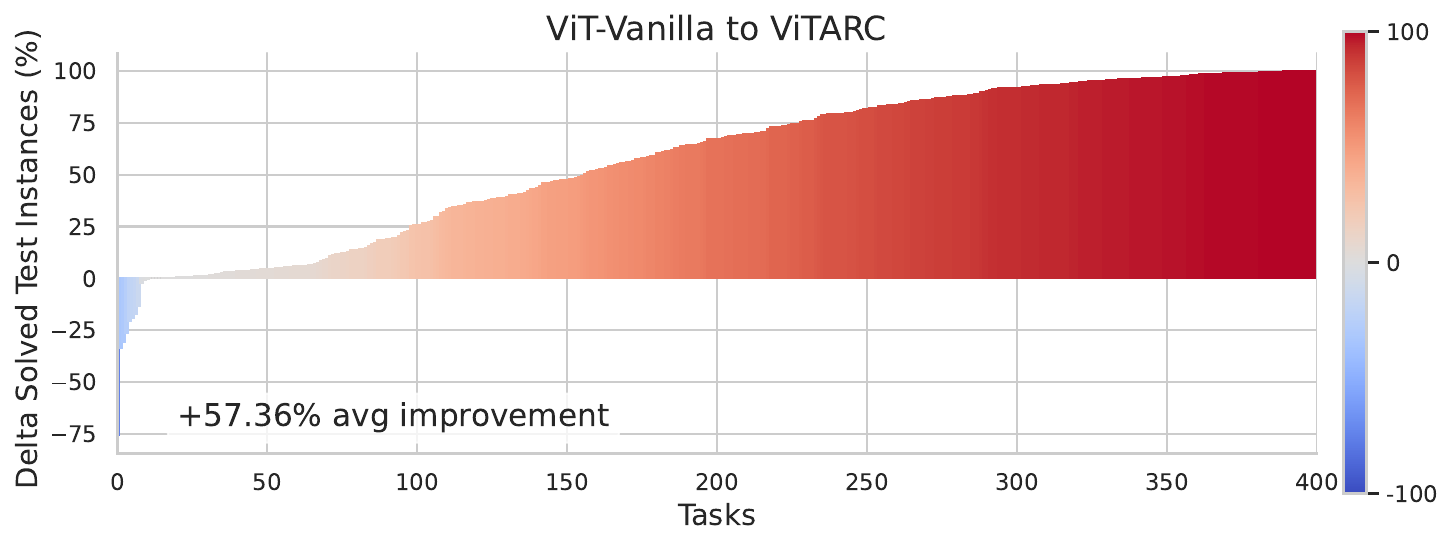}
    \caption{Improvement in percentage of solved test instances per task, from ViT-Vanilla to ViTARC.}
    \label{fig:pertask_full}
\end{figure}

\subsection{Ablation models on sampled 100 tasks}
\label{appendix:substep_ablation}

\begin{table}[htbp]
\caption{Solved test instances (\%) across models on sampled 100 tasks and ablation of sub-steps. The Delta (Mean) column shows the change in the mean solved instances: the ``Border Tokens'' ablation is compared to ViTARC-VT, while the three positional encoding ablations (PEmixer, 2D RPE, and OPE) are compared to ViTARC. Note that the numbers for ViT-Vanilla, ViTARC-VT, and ViTARC differ from the 400-task table as these are based on the 100-task subset.}

\centering
\resizebox{\textwidth}{!}{
\begin{tabular}{l|lllll}
\toprule
\multirow{2}{*}{Model} & \multicolumn{4}{c}{Solved Test Instances (\%)} & \multirow{2}{*}{\textbf{Delta (Mean)}} \\ 
                 & \multicolumn{1}{l}{Mean} & \multicolumn{1}{l}{Median} & \multicolumn{1}{l}{25th Pctl.} & \multicolumn{1}{l}{75th Pctl.} & \\ \midrule
Baseline (ViT-Vanilla)     & \multicolumn{1}{l}{15.98} & \multicolumn{1}{l}{3.65} & \multicolumn{1}{l}{0.10} & 15.90 & \multicolumn{1}{l}{-} \\
\midrule
ViTARC-VT                 & \multicolumn{1}{l}{67.30} & \multicolumn{1}{l}{90.00} & \multicolumn{1}{l}{32.77} & 99.42 & \multicolumn{1}{l}{base} \\
- Border Tokens & \multicolumn{1}{l}{62.71}   & \multicolumn{1}{l}{79.60}     & \multicolumn{1}{l}{28.62}     & 98.80   & \multicolumn{1}{l}{-4.59} \\ 
\midrule
ViTARC (Full Model)        & \multicolumn{1}{l}{78.77} & \multicolumn{1}{l}{95.50} & \multicolumn{1}{l}{78.20} & 99.83 & \multicolumn{1}{l}{base} \\ 
- Positional Encoding Mixer (PEmixer) & \multicolumn{1}{l}{73.03} & \multicolumn{1}{l}{91.25} & \multicolumn{1}{l}{54.90} & 99.05 & \multicolumn{1}{l}{-5.74} \\
- 2D Relative Positional Encoding (2D RPE) & \multicolumn{1}{l}{60.78} & \multicolumn{1}{l}{73.30} & \multicolumn{1}{l}{28.85} & 97.30 & \multicolumn{1}{l}{-17.99} \\
- Object-based Positional Encoding (OPE) & \multicolumn{1}{l}{75.39} & \multicolumn{1}{l}{95.45} & \multicolumn{1}{l}{64.22} & 99.72 & \multicolumn{1}{l}{-3.38} \\ \bottomrule
\end{tabular}
}

\label{tab:sample100_table}
\end{table}

\begingroup
\renewcommand{\arraystretch}{0.9}  

\begin{table}[!htbp]
\vspace{-10mm} 
\centering
\caption{Exact Match Scores for each task on 100 sampled tasks across different models and ablations.}
\label{tab:sampled_100_ablation_table}
\resizebox{\textwidth}{!}{

\begin{tabular}{c!{\vrule width 1pt}c!{\vrule width 1pt}cc!{\vrule width 1pt}ccccc}
\toprule
Task & ViT-Vanilla & ViTARC & $-$BorderTokens & ViTARC & $-$PEmixer & $-$RPE & $-$OPE \\
 & & -VT & & &  & &  \\
\midrule
0ca9ddb6 & 0.00 & 1.00 & 1.00 & 1.00 & 0.27 & 1.00 & 1.00 \\
543a7ed5 & 0.00 & 1.00 & 1.00 & 1.00 & 1.00 & 1.00 & 1.00 \\
952a094c & 0.00 & 1.00 & 0.98 & 1.00 & 0.99 & 1.00 & 0.17 \\
49d1d64f & 0.00 & 1.00 & 1.00 & 1.00 & 1.00 & 1.00 & 1.00 \\
25d487eb & 0.00 & 0.95 & 0.99 & 0.99 & 0.08 & 0.95 & 0.37 \\
d687bc17 & 0.00 & 0.97 & 0.40 & 0.99 & 0.38 & 0.99 & 0.78 \\
67a3c6ac & 0.00 & 1.00 & 0.84 & 0.97 & 0.99 & 1.00 & 1.00 \\
e98196ab & 0.00 & 0.99 & 0.96 & 0.95 & 0.92 & 1.00 & 0.09 \\
8403a5d5 & 0.00 & 1.00 & 0.98 & 0.93 & 0.72 & 0.97 & 0.94 \\
31aa019c & 0.00 & 0.82 & 0.69 & 0.90 & 0.89 & 0.99 & 0.81 \\
ec883f72 & 0.00 & 0.87 & 0.87 & 0.90 & 0.79 & 0.95 & 0.82 \\
b7249182 & 0.00 & 0.74 & 0.61 & 0.88 & 0.81 & 0.90 & 0.32 \\
e76a88a6 & 0.00 & 0.00 & 0.91 & 0.87 & 0.00 & 0.06 & 0.00 \\
3ac3eb23 & 0.00 & 0.71 & 0.71 & 0.87 & 0.85 & 0.87 & 0.57 \\
a64e4611 & 0.00 & 0.98 & 0.97 & 0.87 & 0.90 & 0.99 & 0.99 \\
40853293 & 0.00 & 0.99 & 0.92 & 0.86 & 0.98 & 0.98 & 0.96 \\
b527c5c6 & 0.00 & 0.66 & 0.74 & 0.84 & 0.56 & 0.76 & 0.53 \\
2013d3e2 & 0.00 & 0.95 & 0.92 & 0.84 & 0.11 & 0.94 & 0.94 \\
1caeab9d & 0.00 & 0.42 & 0.78 & 0.82 & 0.48 & 0.58 & 0.36 \\
5521c0d9 & 0.00 & 0.75 & 0.69 & 0.79 & 0.76 & 0.80 & 0.71 \\
6aa20dc0 & 0.00 & 0.33 & 0.52 & 0.63 & 0.38 & 0.51 & 0.23 \\
2dd70a9a & 0.00 & 0.33 & 0.32 & 0.59 & 0.35 & 0.51 & 0.30 \\
5c2c9af4 & 0.00 & 0.51 & 0.40 & 0.46 & 0.53 & 0.53 & 0.31 \\
5daaa586 & 0.00 & 0.17 & 0.48 & 0.43 & 0.22 & 0.37 & 0.12 \\
6e19193c & 0.00 & 0.12 & 0.18 & 0.37 & 0.29 & 0.08 & 0.08 \\
1b60fb0c & 0.00 & 0.14 & 0.17 & 0.28 & 0.06 & 0.12 & 0.04 \\
9aec4887 & 0.00 & 0.02 & 0.11 & 0.19 & 0.01 & 0.03 & 0.00 \\
8a004b2b & 0.00 & 0.02 & 0.10 & 0.18 & 0.02 & 0.11 & 0.00 \\
1f0c79e5 & 0.00 & 0.14 & 0.06 & 0.16 & 0.02 & 0.29 & 0.11 \\
a87f7484 & 0.00 & 0.00 & 0.01 & 0.03 & 0.00 & 0.14 & 0.00 \\
be94b721 & 0.00 & 0.00 & 0.02 & 0.02 & 0.01 & 0.00 & 0.00 \\
c8cbb738 & 0.00 & 0.00 & 0.01 & 0.00 & 0.00 & 0.01 & 0.00 \\
e5062a87 & 0.00 & 0.00 & 0.00 & 0.00 & 0.00 & 0.00 & 0.00 \\
d43fd935 & 0.01 & 0.98 & 0.98 & 1.00 & 0.97 & 0.99 & 0.99 \\
dbc1a6ce & 0.01 & 1.00 & 0.92 & 1.00 & 0.99 & 1.00 & 0.99 \\
dc1df850 & 0.01 & 1.00 & 1.00 & 1.00 & 1.00 & 1.00 & 1.00 \\
dc433765 & 0.01 & 0.91 & 0.92 & 0.96 & 0.68 & 0.97 & 0.94 \\
39a8645d & 0.01 & 0.01 & 0.99 & 0.94 & 0.70 & 0.16 & 0.01 \\
4522001f & 0.01 & 0.22 & 0.62 & 0.88 & 0.74 & 0.79 & 0.76 \\
3f7978a0 & 0.01 & 0.04 & 0.12 & 0.14 & 0.06 & 0.11 & 0.01 \\
d13f3404 & 0.02 & 1.00 & 1.00 & 1.00 & 1.00 & 1.00 & 1.00 \\
913fb3ed & 0.02 & 1.00 & 1.00 & 1.00 & 0.98 & 1.00 & 0.99 \\
94f9d214 & 0.02 & 0.74 & 0.51 & 0.96 & 0.08 & 0.98 & 0.93 \\
228f6490 & 0.02 & 0.03 & 0.06 & 0.06 & 0.04 & 0.04 & 0.02 \\
bdad9b1f & 0.03 & 1.00 & 1.00 & 1.00 & 1.00 & 1.00 & 1.0 \\
eb281b96 & 0.03 & 1.00 & 1.00 & 1.00 & 1.00 & 1.00 & 1.00 \\
6430c8c4 & 0.03 & 0.89 & 0.53 & 0.98 & 0.43 & 0.99 & 0.96 \\
a740d043 & 0.03 & 0.84 & 0.65 & 0.84 & 0.64 & 0.82 & 0.46 \\
d2abd087 & 0.03 & 0.09 & 0.12 & 0.15 & 0.09 & 0.11 & 0.07 \\
5bd6f4ac & 0.04 & 1.00 & 1.00 & 1.00 & 1.00 & 1.00 & 1.00 \\
\bottomrule
\end{tabular}
}
\end{table}

\begin{table}[!htbp]
\vspace{-10mm} 
\centering
\caption{Exact Match Scores for each task on 100 sampled tasks across different models and ablations.}
\resizebox{\textwidth}{!}{

\begin{tabular}{c!{\vrule width 1pt}c!{\vrule width 1pt}cc!{\vrule width 1pt}ccccc}
\toprule
Task & ViT-Vanilla & ViTARC & $-$BorderTokens & ViTARC & $-$PEmixer & $-$RPE & $-$OPE \\
 & & -VT & & &  & &  \\
\midrule
8d5021e8 & 0.04 & 1.00 & 1.00 & 1.00 & 1.00 & 0.96 & 1.00 \\
6150a2bd & 0.04 & 1.00 & 0.57 & 1.00 & 0.69 & 1.00 & 1.00 \\
3af2c5a8 & 0.04 & 1.00 & 1.00 & 0.99 & 1.00 & 1.00 & 0.98 \\
6d0160f0 & 0.04 & 0.03 & 0.93 & 0.97 & 0.02 & 0.98 & 0.56 \\
29ec7d0e & 0.04 & 0.62 & 0.69 & 0.83 & 0.64 & 0.88 & 0.64 \\
760b3cac & 0.04 & 0.66 & 0.60 & 0.64 & 0.11 & 0.78 & 0.47 \\
6c434453 & 0.05 & 0.93 & 0.92 & 0.96 & 0.41 & 0.93 & 0.91 \\
c1d99e64 & 0.05 & 0.99 & 0.92 & 0.95 & 0.90 & 0.95 & 0.96 \\
2dc579da & 0.05 & 0.38 & 0.55 & 0.69 & 0.43 & 0.71 & 0.16 \\
beb8660c & 0.05 & 0.09 & 0.06 & 0.09 & 0.08 & 0.13 & 0.06 \\
7f4411dc & 0.06 & 0.95 & 0.98 & 0.99 & 0.90 & 1.00 & 0.97 \\
32597951 & 0.06 & 0.98 & 0.98 & 0.99 & 0.97 & 1.00 & 0.99 \\
1c786137 & 0.07 & 0.05 & 0.76 & 0.75 & 0.05 & 0.80 & 0.44 \\
d5d6de2d & 0.08 & 0.98 & 1.00 & 1.00 & 0.30 & 0.99 & 0.92 \\
1f642eb9 & 0.08 & 1.00 & 0.92 & 1.00 & 0.89 & 1.00 & 0.98 \\
c444b776 & 0.08 & 0.96 & 0.98 & 0.99 & 0.93 & 0.98 & 0.82 \\
0dfd9992 & 0.09 & 0.67 & 0.82 & 0.84 & 0.73 & 0.83 & 0.66 \\
7837ac64 & 0.09 & 0.82 & 0.85 & 0.82 & 0.85 & 0.79 & 0.60 \\
a61f2674 & 0.10 & 0.75 & 0.71 & 0.84 & 0.84 & 0.86 & 0.54 \\
ce9e57f2 & 0.10 & 0.75 & 0.80 & 0.83 & 0.76 & 0.71 & 0.50 \\
b2862040 & 0.12 & 0.30 & 0.35 & 0.39 & 0.34 & 0.39 & 0.32 \\
d90796e8 & 0.13 & 1.00 & 1.00 & 1.00 & 0.85 & 1.00 & 1.00 \\
42a50994 & 0.14 & 0.98 & 0.97 & 1.00 & 0.82 & 0.99 & 0.24 \\
2bee17df & 0.14 & 0.99 & 1.00 & 1.00 & 0.01 & 1.00 & 0.98 \\
ddf7fa4f & 0.15 & 0.78 & 0.87 & 0.95 & 0.86 & 0.81 & 0.85 \\
7e0986d6 & 0.18 & 0.97 & 1.00 & 1.00 & 1.00 & 0.99 & 0.99 \\
ea786f4a & 0.19 & 0.98 & 0.99 & 0.98 & 0.39 & 0.99 & 0.99 \\
44d8ac46 & 0.20 & 0.59 & 0.70 & 0.86 & 0.79 & 0.74 & 0.63 \\
868de0fa & 0.24 & 1.00 & 1.00 & 1.00 & 1.00 & 0.99 & 1.00 \\
dc0a314f & 0.24 & 0.14 & 0.24 & 0.24 & 0.28 & 0.35 & 0.01 \\
9af7a82c & 0.26 & 0.00 & 0.00 & 0.00 & 0.00 & 0.00 & 0.00 \\
99fa7670 & 0.29 & 1.00 & 1.00 & 1.00 & 0.97 & 1.00 & 1.00 \\
b0c4d837 & 0.29 & 0.21 & 0.91 & 0.97 & 0.21 & 0.93 & 0.13 \\
d89b689b & 0.41 & 0.96 & 0.97 & 0.98 & 0.93 & 0.98 & 0.38 \\
de1cd16c & 0.41 & 0.37 & 0.97 & 0.97 & 0.60 & 0.96 & 0.38 \\
a68b268e & 0.44 & 1.00 & 0.93 & 1.00 & 1.00 & 1.00 & 0.98 \\
d406998b & 0.46 & 0.98 & 1.00 & 1.00 & 0.39 & 1.00 & 0.73 \\
c3f564a4 & 0.52 & 0.94 & 0.94 & 1.00 & 0.88 & 1.00 & 0.92 \\
44f52bb0 & 0.57 & 0.55 & 0.78 & 0.84 & 0.66 & 0.66 & 0.54 \\
ac0a08a4 & 0.61 & 0.99 & 0.98 & 1.00 & 1.00 & 1.00 & 1.00 \\
cce03e0d & 0.64 & 1.00 & 1.00 & 1.00 & 1.00 & 1.00 & 1.00 \\
007bbfb7 & 0.65 & 0.99 & 1.00 & 1.00 & 0.84 & 1.00 & 1.00 \\
91413438 & 0.65 & 0.38 & 0.34 & 0.32 & 0.33 & 0.32 & 0.92 \\
d631b094 & 0.66 & 0.41 & 0.43 & 0.64 & 0.64 & 0.73 & 0.05 \\
445eab21 & 0.86 & 0.92 & 0.97 & 0.96 & 0.92 & 0.92 & 0.90 \\
46f33fce & 0.96 & 1.00 & 1.00 & 1.00 & 0.84 & 1.00 & 1.00 \\
5582e5ca & 0.98 & 0.95 & 1.00 & 0.99 & 0.98 & 0.97 & 0.96 \\
c8f0f002 & 0.99 & 1.00 & 1.00 & 1.00 & 1.00 & 1.00 & 1.00 \\
995c5fa3 & 1.00 & 0.00 & 1.00 & 1.00 & 1.00 & 0.02 & 1.00 \\
6e02f1e3 & 1.00 & 1.00 & 0.89 & 1.00 & 1.00 & 1.00 & 0.96 \\
\bottomrule
\end{tabular}
}
\label{tab:sampled_100_ablation_table}
\end{table}

\endgroup

\newpage

\section{Additional Ablation on \textit{PEmixer}, \textit{2DRPE}, and \textit{OPE}}
\label{appendix:ablations_pemixer_2drpe_ope}

In Section~\ref{sec:PositionalEnhancements}, we introduced the full \method{} (ViTARC) model, incorporating 
\textit{PEmixer}, \textit{2DRPE}, and \textit{OPE} on top of the \textit{ViTARC-VT} baseline. 
Here, we present additional ablation results demonstrating how much each module \emph{gains} when it is \textbf{added back} 
to the ablated model. Concretely, we remove each module from the full model and measure the drop in performance 
(i.e., the ``gain'' that module provided). We continue to analyze the same 100-task subset introduced in 
Appendix~\ref{appendix:substep_ablation}. 

The tables below highlight tasks that experience a \emph{significant performance gain} 
(>\,0.40 in exact-match score) from adding the module back, or a \emph{moderate loss} (>\,0.05) 
when the module is present. These latter cases are tasks that might do slightly better without 
the corresponding module. 

\subsection{Ablation Tables for Each Module}

\paragraph{1) \textbf{PEmixer.}}
Tables~\ref{tab:ablate_pemixer_gain} and \ref{tab:ablate_pemixer_loss} list tasks 
that \emph{gain or lose} more than certain thresholds when \textit{PEmixer} is added back. 
A gain above 0.40 in exact-match is \emph{significant}, while a drop over 0.05 is considered 
a \emph{moderate loss}.

\begin{table}[h]
    \centering
    \caption{Tasks gaining more than 0.40 exact-match by adding PEmixer. 
    (These were significantly worse when PEmixer was removed.)}
    \label{tab:ablate_pemixer_gain}
    \begin{tabular}{r|rrr}
    \toprule
    \textbf{task\_id} & \textbf{exact\_match\_full} & \textbf{exact\_match\_noPEmixer} & \textbf{diff\_noPEmixer} \\
    \midrule
    d687bc17 & 0.989 & 0.402 & 0.587\\
    94f9d214 & 0.956 & 0.511 & 0.445\\
    6430c8c4 & 0.977 & 0.534 & 0.443\\
    6150a2bd & 1.000 & 0.572 & 0.428\\
    \bottomrule
    \end{tabular}
\end{table}

\begin{table}[h]
    \centering
    \caption{Tasks losing more than 0.05 exact-match by adding PEmixer.
    (These slightly improve if PEmixer is removed.)}
    \label{tab:ablate_pemixer_loss}
    \begin{tabular}{r|rrr}
    \toprule
    \textbf{task\_id} & \textbf{exact\_match\_full} & \textbf{exact\_match\_noPEmixer} & \textbf{diff\_noPEmixer} \\
    \midrule
    a64e4611 & 0.871 & 0.967 & -0.096\\
    2013d3e2 & 0.842 & 0.917 & -0.075\\
    39a8645d & 0.936 & 0.992 & -0.056\\
    40853293 & 0.862 & 0.916 & -0.054\\
    \bottomrule
    \end{tabular}
\end{table}

\paragraph{2) \textbf{2DRPE.}}
Tables~\ref{tab:ablate_rpe_gain} and \ref{tab:ablate_rpe_loss} show tasks that gain or lose 
significantly when 2DRPE is present.

\begin{table}[h]
    \centering
    \caption{Tasks gaining more than 0.40 exact-match by adding 2DRPE.}
    \label{tab:ablate_rpe_gain}
    \begin{tabular}{r|rrr}
    \toprule
    \textbf{task\_id} & \textbf{exact\_match\_full} & \textbf{exact\_match\_noRPE} & \textbf{diff\_noRPE}\\
    \midrule
    2bee17df & 0.998 & 0.013 & 0.985\\
    6d0160f0 & 0.970 & 0.019 & 0.951\\
    25d487eb & 0.992 & 0.075 & 0.917\\
    94f9d214 & 0.956 & 0.081 & 0.875\\
    e76a88a6 & 0.867 & 0.000 & 0.867\\
    b0c4d837 & 0.973 & 0.211 & 0.762\\
    2013d3e2 & 0.842 & 0.106 & 0.736\\
    0ca9ddb6 & 1.000 & 0.272 & 0.728\\
    1c786137 & 0.749 & 0.046 & 0.703\\
    d5d6de2d & 1.000 & 0.299 & 0.701\\
    d687bc17 & 0.989 & 0.378 & 0.611\\
    d406998b & 1.000 & 0.391 & 0.609\\
    ea786f4a & 0.985 & 0.388 & 0.597\\
    6c434453 & 0.957 & 0.409 & 0.548\\
    6430c8c4 & 0.977 & 0.432 & 0.545\\
    760b3cac & 0.642 & 0.108 & 0.534\\
    \bottomrule
    \end{tabular}
\end{table}

\begin{table}[h]
    \centering
    \caption{Tasks losing more than 0.05 exact-match by adding 2DRPE.}
    \label{tab:ablate_rpe_loss}
    \begin{tabular}{r|rrr}
    \toprule
    \textbf{task\_id} & \textbf{exact\_match\_full} & \textbf{exact\_match\_noRPE} & \textbf{diff\_noRPE}\\
    \midrule
    40853293 & 0.862 & 0.983 & -0.121\\
    5c2c9af4 & 0.464 & 0.531 & -0.067\\
    \bottomrule
    \end{tabular}
\end{table}

\paragraph{3) \textbf{OPE.}}
Lastly, Tables~\ref{tab:ablate_ope_gain} and \ref{tab:ablate_ope_loss} detail tasks 
that gain or lose when OPE is present.

\begin{table}[h]
    \centering
    \caption{Tasks gaining more than 0.40 exact-match by adding OPE.}
    \label{tab:ablate_ope_gain}
    \begin{tabular}{r|rrr}
    \toprule
    \textbf{task\_id} & \textbf{exact\_match\_full} & \textbf{exact\_match\_noOPE} & \textbf{diff\_noOPE}\\
    \midrule
    995c5fa3 & 1.000 & 0.019 & 0.981\\
    e76a88a6 & 0.867 & 0.059 & 0.808\\
    39a8645d & 0.936 & 0.156 & 0.780\\
    \bottomrule
    \end{tabular}
\end{table}

\begin{table}[h]
    \centering
    \caption{Tasks losing more than 0.05 exact-match by adding OPE.}
    \label{tab:ablate_ope_loss}
    \begin{tabular}{r|rrr}
    \toprule
    \textbf{task\_id} & \textbf{exact\_match\_full} & \textbf{exact\_match\_noOPE} & \textbf{diff\_noOPE}\\
    \midrule
    760b3cac & 0.642 & 0.778 & -0.136\\
    1f0c79e5 & 0.163 & 0.293 & -0.130\\
    a64e4611 & 0.871 & 0.990 & -0.119\\
    a87f7484 & 0.029 & 0.144 & -0.115\\
    40853293 & 0.862 & 0.975 & -0.113\\
    dc0a314f & 0.241 & 0.349 & -0.108\\
    2013d3e2 & 0.842 & 0.941 & -0.099\\
    31aa019c & 0.897 & 0.990 & -0.093\\
    d631b094 & 0.641 & 0.729 & -0.088\\
    5c2c9af4 & 0.464 & 0.527 & -0.063\\
    1c786137 & 0.749 & 0.802 & -0.053\\
    29ec7d0e & 0.831 & 0.882 & -0.051\\
    \bottomrule
    \end{tabular}
\end{table}

\subsection{Overall Gains vs.\ Losses}
\begin{table}[h]
    \centering
    \resizebox{\textwidth}{!}{%
    \begin{tabular}{l|rrrrrrr}
    \toprule
    \textbf{Module} & \textbf{mean\_gain} & \textbf{count\_gain} & \textbf{max\_gain} 
                    & \textbf{mean\_loss} & \textbf{count\_loss} & \textbf{max\_loss}
                    & \textbf{overall\_mean\_diff}\\
    \midrule
    RPE      & 0.2351 & 78 & 0.985 & -0.0297 & 12 & -0.121 & 0.17985 \\
    OPE      & 0.1169 & 42 & 0.981 & -0.0404 & 38 & -0.136 & 0.03372 \\
    PEmixer  & 0.0980 & 64 & 0.587 & -0.0233 & 23 & -0.096 & 0.05734 \\
    \bottomrule
    \end{tabular}%
    }
    \caption{Mean gains/losses across 100 tasks upon \emph{adding back} each module into the ablated model.
    A positive ``mean\_gain'' indicates how much that module typically boosts performance (with 
    \texttt{max\_gain} showing the single largest improvement), while ``mean\_loss'' 
    and \texttt{max\_loss} show the severity of the drop for tasks that prefer it removed.}
    \label{tab:summary_ablation}
\end{table}

\paragraph{Further Insights and Case Analysis.}
From Table~\ref{tab:summary_ablation}, we see that the \emph{gains} from each module
are considerably larger than any \emph{losses}:
\begin{enumerate}
    \item The largest gains approach +98.5\% in exact-match score,
          whereas the most severe losses rarely exceed -13.6\%.
    \item Both the counts (\texttt{count\_gain} vs.\ \texttt{count\_loss})
          and the mean differences (\texttt{mean\_gain} vs.\ \texttt{mean\_loss})
          reinforce that each module brings a net positive impact.
\end{enumerate}

Among the three modules, \textbf{2DRPE} provides the highest average improvement,
likely because it offers a more general mechanism for differentiating
nearby same-color pixels and therefore detecting \textbf{multi-color objects}—both crucial needs in ARC tasks.

\begin{figure}[h]
    \centering
    \includegraphics[width=0.99\linewidth]{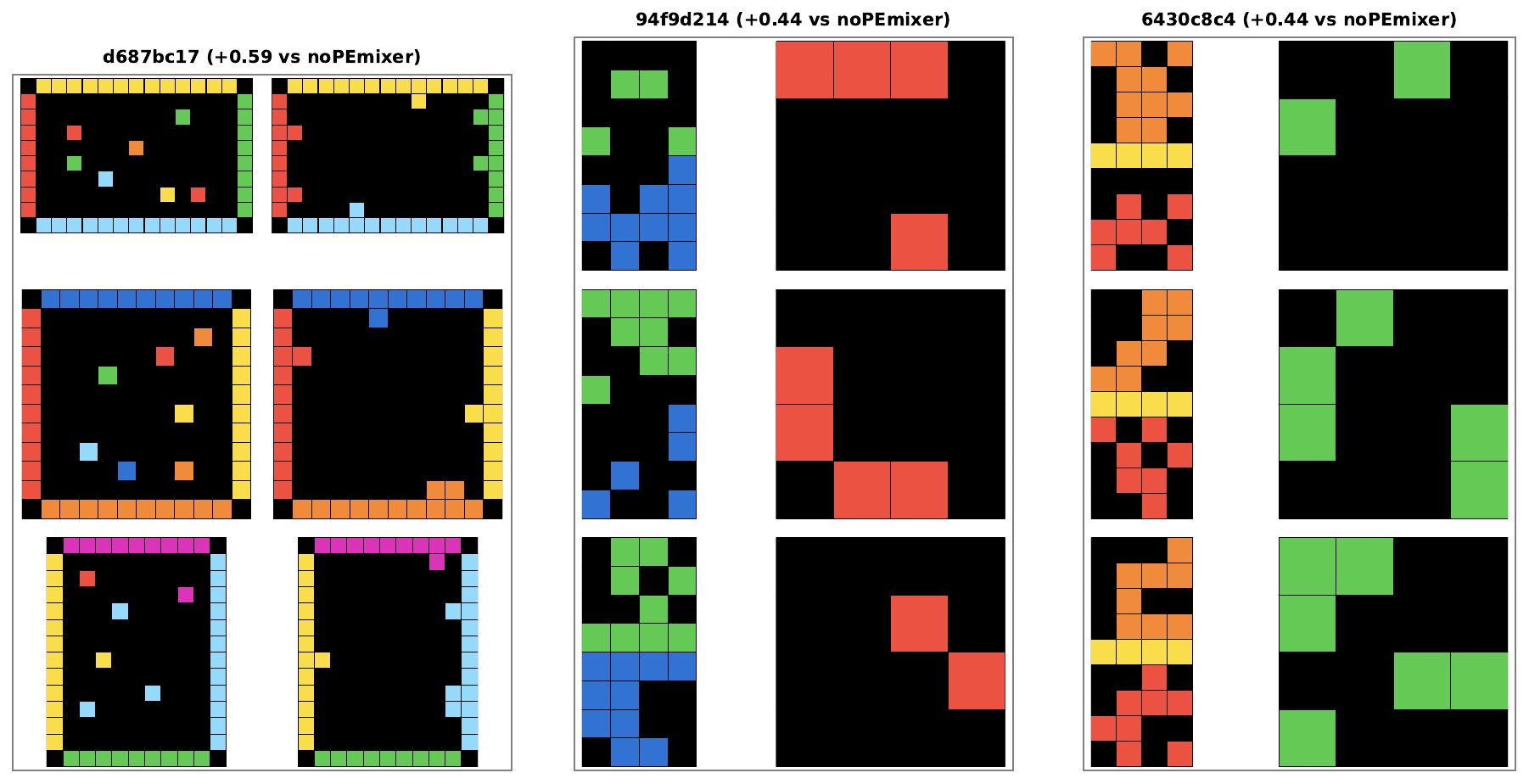}
    \caption{Example tasks with largest gains from adding PEmixer.}
    \label{fig:top3_PEmixer_gain}
\end{figure}

\begin{figure}[h]
    \centering
    \includegraphics[width=0.99\linewidth]{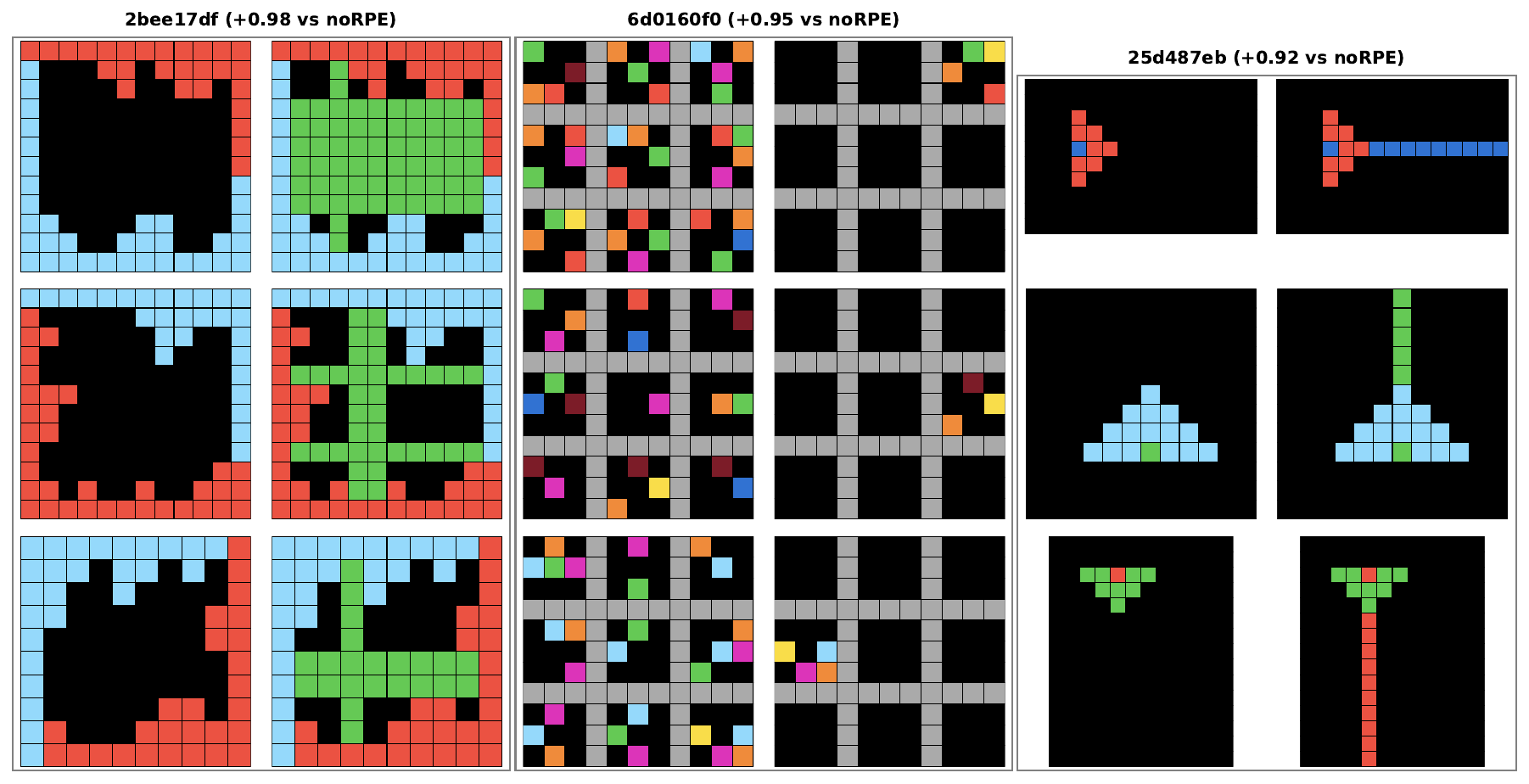}
    \caption{Example tasks with largest gains from adding 2DRPE.}
    \label{fig:top3_RPE_gain}
\end{figure}

\begin{figure}[h]
    \centering
    \includegraphics[width=0.99\linewidth]{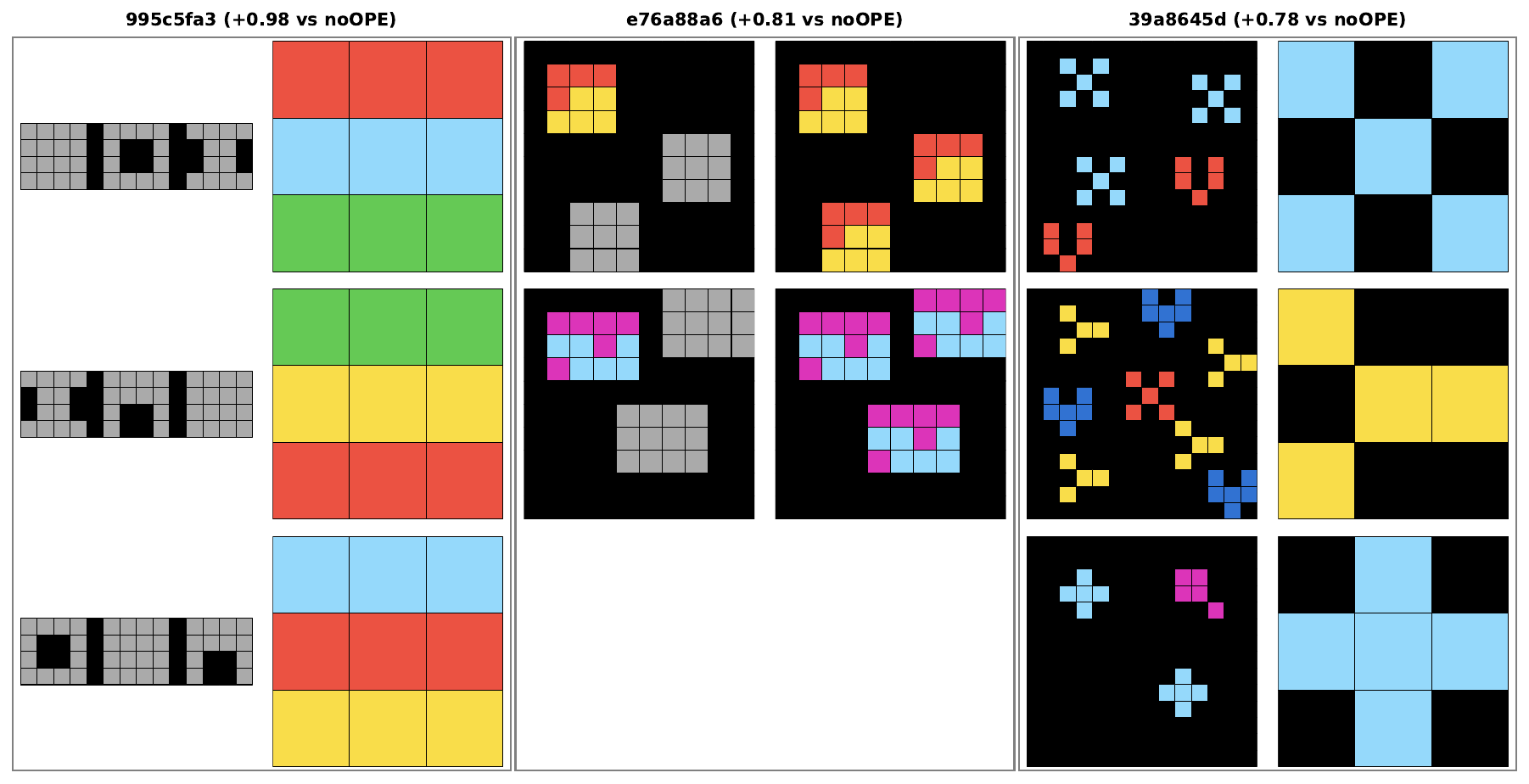}
    \caption{Example tasks with largest gains from adding OPE.}
    \label{fig:top3_OPE_gain}
\end{figure}

We further illustrate these findings by plotting the top-3 performance gains 
for each module (Figures~\ref{fig:top3_PEmixer_gain}, \ref{fig:top3_RPE_gain}, and \ref{fig:top3_OPE_gain}). 
These concrete examples make the benefits of each component more intuitive:

\begin{enumerate}
    \item \textbf{PEmixer:} As hypothesized, it boosts tasks where spatial positioning outweighs
    color cues. In tasks like \textbf{94f9d214} and \textbf{6430c8c4}, for example, the solution involves overlapping
    two separate subgrids or identifying uncolored cells. Color simply marks different boundaries,
    but \emph{where} each pixel belongs is the more relevant factor. PEmixer increases the
    relative importance of positional embeddings, helping the model solve these puzzles.

    \item \textbf{2DRPE:} Compared to standard APE, 2DRPE precisely distinguishes pixels of the same color
    that lie close to each other. This local offset information is vital for determining the
    boundaries of multi-color objects. We observe tasks gaining over 90\% in exact-match 
    purely from this additional spatial bias, underscoring how 2DRPE refines pixel-level 
    discrimination.

    \item \textbf{OPE:} While some tasks appear to revolve around multi-color objects, they remain
    challenging for pure RPE if no direct correspondence between input and output grids 
    indicates how fragments should be combined. For instance, in \textbf{39a8645d}, multiple 
    diagonally connected red pixels form one logical object, but RPE alone does not clarify 
    if it is one, two, or three objects. By contrast, humans naturally assume these fragments 
    unite into a single entity—an assumption captured by OPE’s “objectness” prior.
    \footnote{In our experiments, we rely on OpenCV contours for demonstration, but more powerful 
    segmentors like SAM could embed a human-like objectness bias learned from real-world data.}
\end{enumerate}

Overall, these examples confirm that \textit{PEmixer}, \textit{2DRPE}, and \textit{OPE} target distinct
aspects of spatial reasoning, and each significantly boosts performance on tasks
aligned with that module’s strengths. Although a small subset of tasks exhibit minor
improvements when a module is removed, the majority see large, tangible benefits
from including all three components in \method{}.

\FloatBarrier 

\subsection{Case Study: Additive Ablation on Task \texttt{f25fbde4}}
\label{appendix:case_study_f25fbde4}

While Figure~\ref{fig:cross_attn} (in the main text) motivated the need for stronger positional cues on a different ARC task, we here provide a similar \emph{additive ablation} analysis for a representative case, \texttt{f25fbde4}. Table~\ref{tab:f25fbde4_ablation} shows the exact-match scores of progressively adding modules (\textit{PEmixer}, \textit{2DRPE}, \textit{OPE}) on top of \texttt{ViTARC-VT} (i.e., Visual Tokens + 2D APE). We see a near-monotonic improvement from individual modules to partial combinations, and finally the full \method{} (PEmixer + 2DRPE + OPE) reaches 0.96 exact-match.

\begin{table}[h]
\centering
\caption{Additive ablation on task \texttt{f25fbde4}. Each row adds one or more modules to \texttt{ViTARC-VT}.}
\vspace{4pt}
\label{tab:f25fbde4_ablation}
\begin{tabular}{l|c}
\toprule
\textbf{Model Variant} & \textbf{Exact-Match} \\
\midrule
\texttt{ViT-Vanilla}                   & 0.03 \\
\texttt{ViTARC-VT}                     & 0.02 \\
\quad + \textit{PEmixer}               & 0.014 \\
\quad + \textit{2DRPE}                 & 0.27 \\
\quad + \textit{OPE}                   & 0.656 \\
\quad + \textit{PEmixer + 2DRPE}       & 0.269 \\
\quad + \textit{2DRPE + OPE}           & 0.924 \\
\quad + \textit{PEmixer + OPE}         & 0.871 \\
\textbf{ViTARC} (\textit{PEmixer + 2DRPE + OPE}) & \textbf{0.96} \\
\bottomrule
\end{tabular}
\end{table}

\paragraph{Cross-Attention Heatmap Comparisons.}
To visualize why each component helps, we plot cross-attention head heatmaps for four intermediate variants:
\begin{itemize}
    \item \(\texttt{ViTARC-VT}\) \quad (score = 0.02)
    \item \(\texttt{ViTARC-VT + PEmixer}\) \quad (score = 0.014)
    \item \(\texttt{ViTARC-VT + 2DRPE}\) \quad (score = 0.27)
    \item \(\texttt{ViTARC-VT + OPE}\) \quad (score = 0.656)
\end{itemize}

\begin{figure}[h]
\centering
\includegraphics[width=0.88\linewidth]{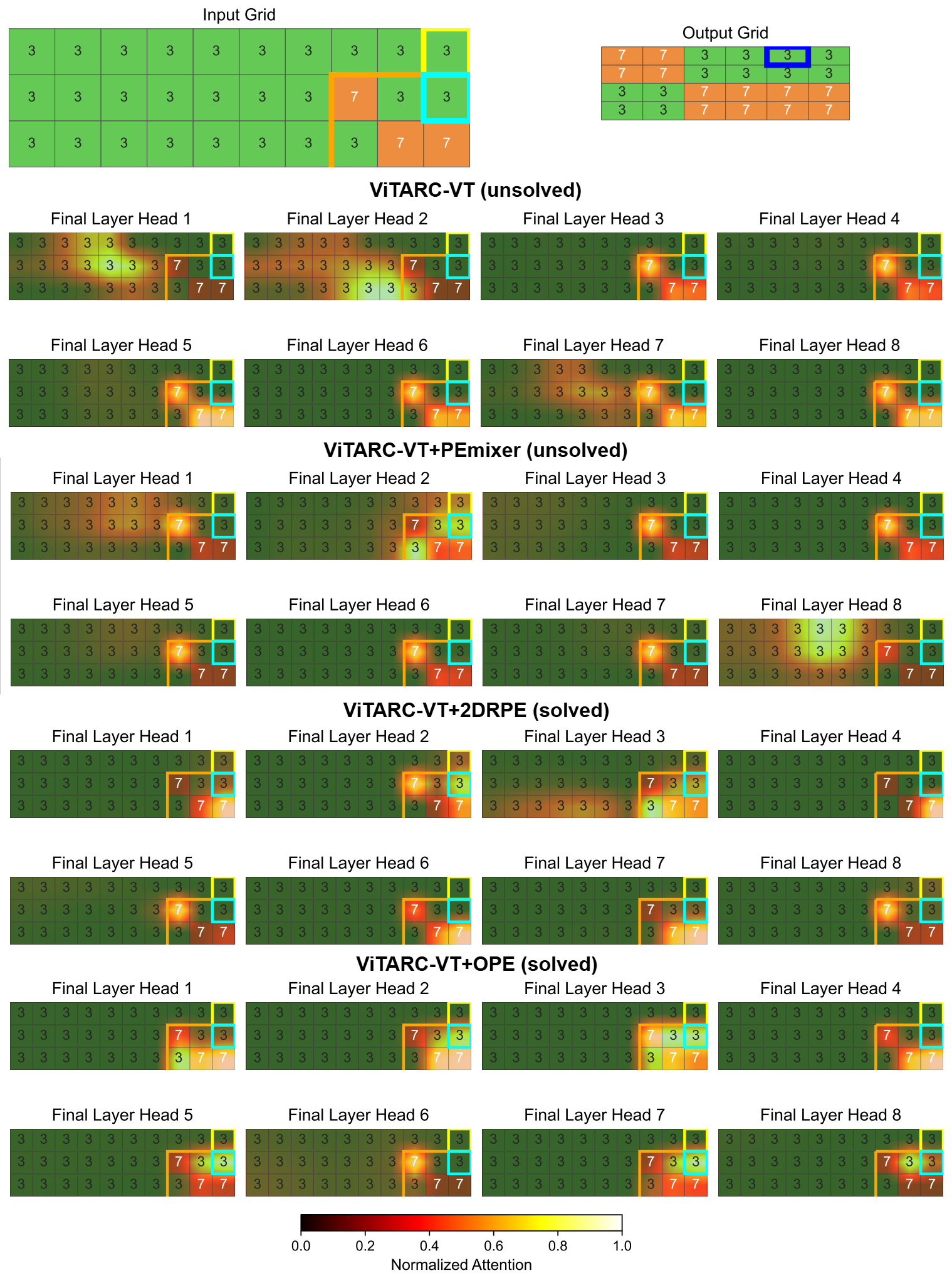}
\caption{\textbf{Cross-attention thermal heatmaps for four partial ViTARC variants on task \texttt{f25fbde4}.}
Each heatmap is taken at the final decoder step for a critical pixel (dark blue box).
The goal is to identify a subgrid (orange bounding box), yet when relying solely on
PEmixer, the model does not properly isolate this region (score~0.014).
Moreover, the color-similar pixel inside the cyan box (within the subgrid) and the pixel
in the yellow box (outside the subgrid) remain indistinguishable without 2DRPE or OPE
(scores~0.27 and~0.656, respectively), illustrating why these additional positional
biases are crucial for accurate attention separation.}
\label{fig:f25fbde4_crs_attn}
\end{figure}

Figure~\ref{fig:f25fbde4_crs_attn} confirms that no single component alone can deliver full performance on this instance. 
When we add only \textit{PEmixer} to \texttt{ViTARC-VT}, the exact-match remains low (0.014): as predicted, simply 
boosting the default APE embeddings does not help the model separate nearby pixels of the same color. By contrast, 
both \texttt{ViTARC-VT + 2DRPE} and \texttt{ViTARC-VT + OPE} correctly solve the example; the heatmaps show that 
2DRPE provides a partial boundary distinction for the subgrid (score 0.27), while OPE more clearly isolates the 
intended region (score 0.656) by injecting objectness priors.

Combining any two of \{\textit{PEmixer, 2DRPE, OPE}\} yields a near-monotonic improvement over a single addition, 
reflecting the synergy among the modules. For instance:
\begin{itemize}
    \item \(\texttt{ViTARC-VT + PEmixer + OPE}\) jumps to 0.871, which is +0.215 over 
          \(\texttt{ViTARC-VT + OPE}\) (0.656), demonstrating that emphasizing OPE’s object boundaries 
          (via PEmixer) can resolve subtle spatial ambiguities.
    \item \(\texttt{ViTARC-VT + 2DRPE + OPE}\) further climbs to 0.924, reflecting how 2DRPE and OPE 
          offer \emph{distinct} inductive biases: one for local pixel differentials, the other for objectness grouping.
\end{itemize}
Finally, \(\texttt{ViTARC}\) integrates all three modules, achieving 0.96 and underscoring that each component 
helps address a separate aspect of the ARC’s pixel-level reasoning challenge.

\end{document}

%% file: math_commands.tex
\usepackage{amsmath,amsfonts,bm}

\def\eqref#1{equation~\ref{#1}}

\def\1{\bm{1}}

\def\mW{{\bm{W}}}

\DeclareMathAlphabet{\mathsfit}{\encodingdefault}{\sfdefault}{m}{sl}
\SetMathAlphabet{\mathsfit}{bold}{\encodingdefault}{\sfdefault}{bx}{n}

